\documentclass[twoside]{article}
\usepackage[accepted]{aistats2022}

\usepackage{microtype}
\usepackage{graphicx}
\usepackage{booktabs} 
\usepackage{sidecap}
\usepackage{pifont}

\usepackage{url}
\usepackage{graphbox}
\usepackage[export]{adjustbox}
\usepackage{microtype}
\usepackage{graphicx}
\usepackage{booktabs} 
\usepackage[table]{xcolor}
\usepackage[T1]{fontenc}    
\usepackage{xurl}
\usepackage{booktabs}       
\usepackage{amsfonts}       
\usepackage{nicefrac}       
\usepackage{microtype}      
\usepackage{amsmath,amssymb,amsthm}
\usepackage{mathtools}  
\usepackage{mathrsfs}
\usepackage{thmtools}
\usepackage{thm-restate}
\usepackage{comment}
\usepackage{parskip}
\usepackage{dblfloatfix}    
\usepackage{siunitx}
\usepackage{etoolbox}
\usepackage{enumitem}
\usepackage{graphicx}
\usepackage{subcaption}
\usepackage[textsize=scriptsize,textwidth=2.2cm]{todonotes}
\usepackage{xcolor} 
\usepackage{tabularx}
\usepackage{longtable} 
\usepackage{supertabular}
\usepackage{wrapfig}
\usepackage{algorithm}
\usepackage{algorithmic}

\newtoggle{showtodos}
\togglefalse{showtodos}

\usepackage{hyperref}

\usepackage{xr}
\makeatletter
\newcommand*{\addFileDependency}[1]{
  \typeout{(#1)}
  \@addtofilelist{#1}
  \IfFileExists{#1}{}{\typeout{No file #1.}}
}
\makeatother

\newcommand*{\myexternaldocument}[1]{
    \externaldocument{#1}
    \addFileDependency{#1.tex}
    \addFileDependency{#1.aux}
}

\myexternaldocument{supplement}
\newcommand{\Eop}{\ensuremath{\mathop{\mathbb{E}}}}
\newcommand{\E}{\Eop}

\newcommand{\Beta}{\ensuremath{\text{Beta}}}
\newcommand{\Uniform}{\ensuremath{\text{Uniform}}}

\DeclareMathOperator*{\argmin}{arg\,min}

\newcommand{\emmonsweep}{\ensuremath{\text{ECE}^\textsc{em}_\textsc{sweep}}}
\newcommand{\emdebias}{\ensuremath{\text{ECE}^\textsc{em}_\textsc{debias}}}
\newcommand{\ewdebias}{\ensuremath{\text{ECE}^\textsc{ew}_\textsc{debias}}}
\newcommand{\kde}{\ensuremath{\text{KDE}}}
\newcommand{\ewece}{\ensuremath{\text{ECE}^\textsc{ew}_\textsc{bin}}}

\newcommand{\EM}{{\textsc{em}}}
\newcommand{\EW}{{\textsc{ew}}}

\newcommand{\ecebin}{\ensuremath{{\text{ECE}_{\textsc{bin}}}}}
\newcommand{\ecedebias}{\ensuremath{{\text{ECE}_{\textsc{debias}}}}}
\newcommand{\celb}{\ensuremath{{\text{ECE}_{\textsc{lb}}}}}
\newcommand{\cesweep}{\ensuremath{{\text{ECE}_{\textsc{sweep}}}}}
\newcommand{\BBC}{{BBC}}

\newcommand{\ceneighborhood}{\ensuremath{{\text{ECE}_{\mathcal{N}}}}}

\newcommand{\truecurve}{\ensuremath{\E_Y[Y \mid f(X)=c]}}
\newcommand{\truecurvex}{\ensuremath{\E_Y[Y \mid f(X)]}}
\newcommand{\truece}{\ensuremath{\text{true calibration error}}}
\newcommand{\tce}{\ensuremath{\text{TCE}}}

\newcolumntype{L}[1]{>{\raggedright\arraybackslash}p{#1}}
\newcolumntype{C}[1]{>{\centering\arraybackslash}p{#1}}
\newcolumntype{R}[1]{>{\raggedleft\arraybackslash}p{#1}}

\iftoggle{showtodos}{
\newcommand{\becca}[1]{\todo[color=green!40]{Becca: #1}}
\newcommand{\nick}[1]{\todo[color=blue!40]{Nick: #1}}
\newcommand{\mike}[1]{\todo[color=red!40]{Mike: #1}}
\newcommand{\jon}[1]{\todo[color=purple!40]{Jon: #1}}
}{
\newcommand{\becca}[1]{}
\newcommand{\nick}[1]{}
\newcommand{\jon}[1]{}
\newcommand{\mike}[1]{}
}

\usepackage[round,sort]{natbib}

\begin{document}

\twocolumn[
\aistatstitle{Mitigating Bias in Calibration Error Estimation}
\aistatsauthor{Rebecca Roelofs \And Nicholas Cain \And Jonathon Shlens \And Michael C. Mozer}
\aistatsaddress{Google Research \And Google Research \And Google Research \And Google Research}
]

\begin{abstract}

For an AI system to be reliable, the confidence it expresses in its decisions must match its accuracy. To assess the degree of match, examples are typically binned by confidence and the per-bin mean confidence and accuracy are compared. Most research in calibration focuses on techniques to reduce this empirical measure of calibration error, \ecebin{}. We instead focus on assessing statistical bias in this empirical measure, and we identify better estimators. We propose a framework through which we can compute the bias of a particular estimator for an evaluation data set of a given size. The framework involves  synthesizing model outputs that have the same statistics as common neural architectures on popular data sets. We find that binning-based estimators with bins of equal mass (number of instances) have lower bias than estimators with bins of equal width. Our results indicate two reliable calibration-error estimators: the debiased estimator 
\citep{brocker2012,ferro2012} and a method we propose, \cesweep{}, which uses equal-mass bins and chooses the number of bins to be as large as possible while preserving monotonicity in the calibration function. With these estimators, we observe improvements in the effectiveness of recalibration methods and in the detection of model miscalibration. 
\end{abstract}

\section{INTRODUCTION}
\label{sec:intro}
Machine learning models are increasingly deployed in high-stakes settings like self-driving cars \citep{sun2020scalability,geiger2013vision,caesar2020nuscenes} and medical diagnosis \citep{esteva2019guide,gulshan2016development,esteva2017dermatologist} where it is critical to recognize when a model is likely to be incorrect.  Unfortunately, models often fail in unexpected and poorly understood ways, hindering our ability to interpret and trust such systems \citep{recht2019imagenet, biggio2017wild, intriguing, hendrycks2018benchmarking, azulay2018deep}. 
To address these issues, calibration is used to ensure that a model produces confidence scores that reflect its ground truth likelihood of being correct \citep{zadrozny2001obtaining, zadrozny2002transforming, platt1999probabilistic}.

To obtain an estimate of the calibration error, or ECE\footnote{\citet{naeini2015obtaining} introduce ECE as an acronym for \textit{Expected} Calibration Error.  However, ECE is not a proper expectation whereas the \truece{} is computed under an expectation. To resolve this confusion, we prefer to read ECE as \textit{Estimated} Calibration Error.}, the standard procedure partitions the model confidence scores into bins and compares the model's predicted accuracy to its empirical accuracy within each bin \citep{guo2017calibration, naeini2015obtaining}. We refer to this specific metric as \ecebin{}. Recent work observed that the calculation of \ecebin{} is sensitive to implementation \citep{kumar2019verified,nixon2019measuring}.
Fundamentally, a key confounding factor is \emph{statistical bias},
the difference between the expected \ecebin{} and the true calibration 
error (\tce{}). Because bias is largely unexplored in the literature,
its magnitude and sign is unknown, 
as is its dependence on hyperparameters of the \ecebin{} estimator (e.g., number 
of bins, how bins are formed).
We explain our reasons for focusing on estimator bias and not variance in Section \ref{sec:simulation_setup}.


Bias in \ecebin{} measurement has two real world consequences. First, the measurement of calibration error on a given model may be systematically incorrect. Thus, our understanding of how well a model knows whether it is correct may be poor, and may not be accurately captured by naively reporting \ecebin{}. Second, many techniques have been developed to minimize the calibration error, such as post-hoc recalibration techniques \citep{guo2017calibration,zadrozny2001obtaining,zadrozny2002transforming} and, more recently, calibration-sensitive training objectives \citep{karandikar2021,kumar2018trainable,mukhoti2020calibrating,Krishnan2020ImprovingMC,lin2018focal}. 
Given that the selection of the training objectives and the justification of a recalibration technique is predicated on the measurement of the calibration error, reliance on an inaccurate estimator may lead to a suboptimal choice.

\begin{SCfigure}[10][t]
  \newcommand{\ppwm}{0.49}
  \centering
    \begin{subfigure}{\ppwm\textwidth}
    \includegraphics[width=\ppwm\linewidth]{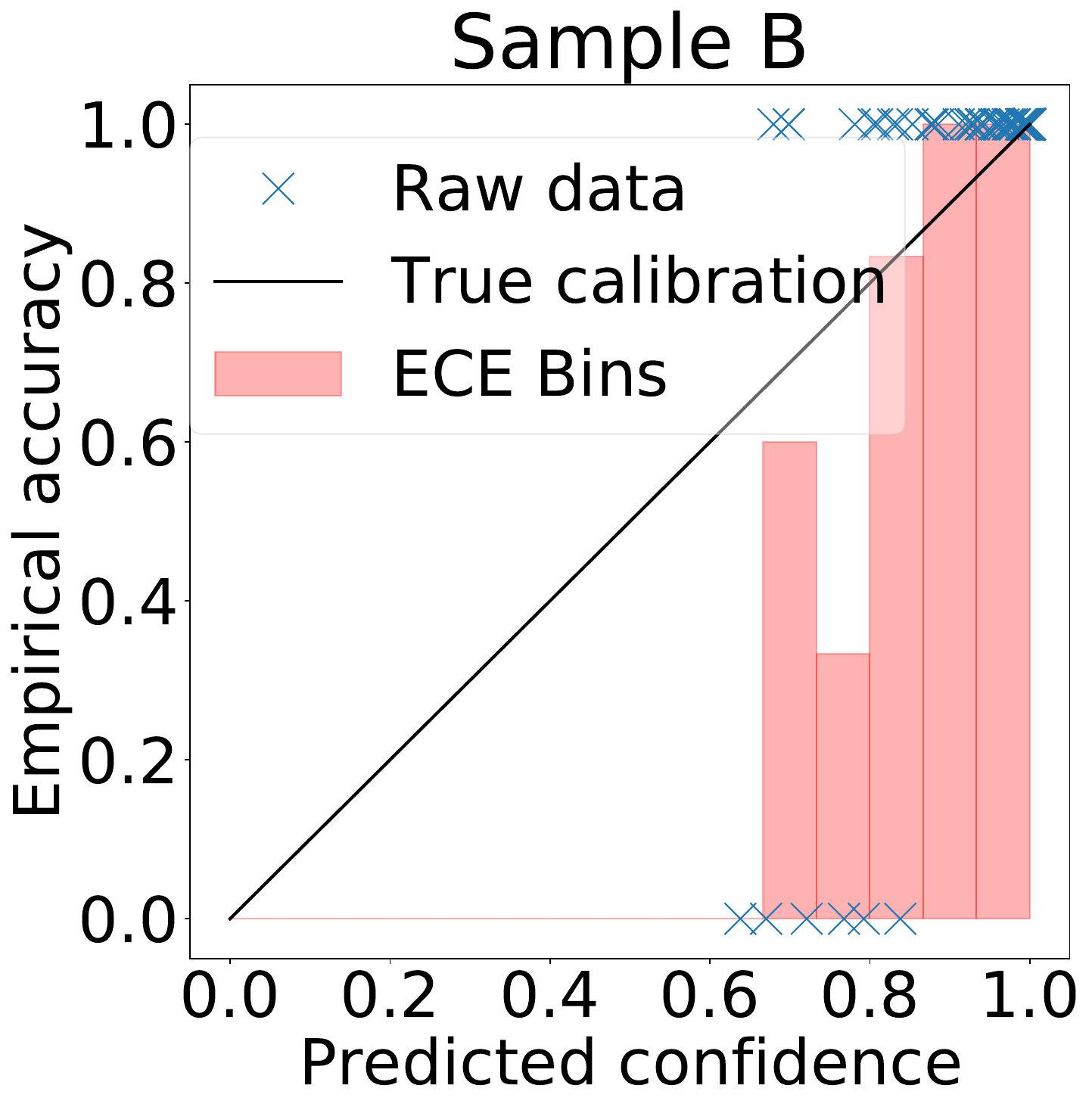}
    \includegraphics[width=\ppwm\linewidth]{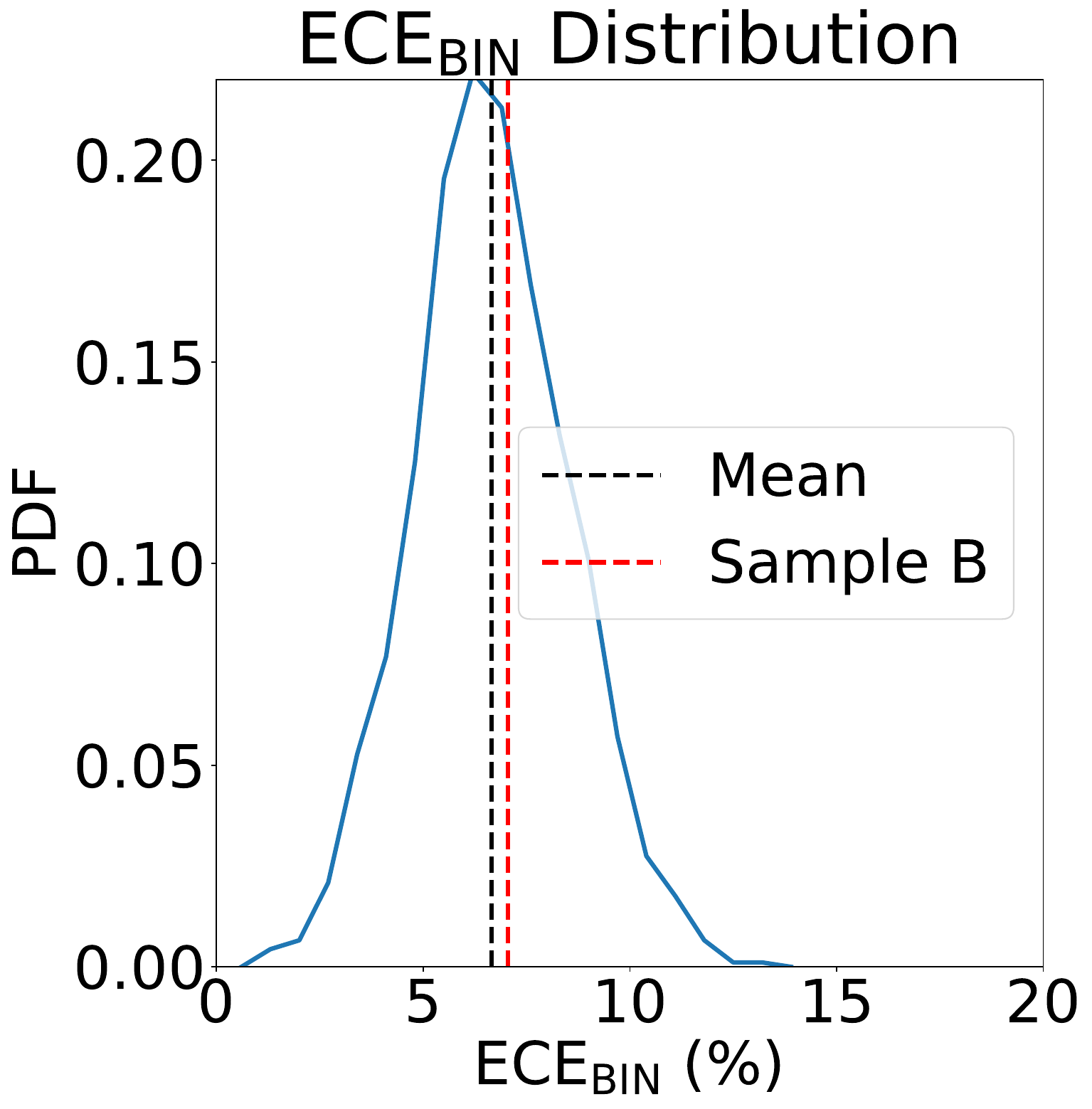}

\vspace{10pt}  
\caption{ 
Figure \ref{fig:intro}: \ecebin{} exhibits large bias for perfectly calibrated models. We simulate data from a perfectly calibrated model with confidence scores fit to ResNet-110 CIFAR-10 output \citep{nn_markus, he2016deep} and measure \ecebin{} using 15 equal-width-spaced bins.
The left panel shows a reliability diagram for a sample of size $n = 200$ (named Sample B);
the right panel has a distribution of \ecebin{} scores computed across $10^6$  independent simulations.
\textit{Even though the model is perfectly calibrated, \ecebin{} systematically predicts large calibration errors.} 
  }
  \vspace{-10pt} 
  \end{subfigure}
  \label{fig:intro}
\end{SCfigure}
\newcommand{\y}{\textcolor{green}{\ding{51}}}
\newcommand{\n}{\textcolor{red}{\ding{55}}}
\begin{table*}[t]
    \centering
    \begin{tabular}{|c|cccc|cccc|cc|}
         \bottomrule
        & \multicolumn{4}{c|}{\textbf{CIFAR-10}} & \multicolumn{4}{c|}{\textbf{CIFAR-100}} & \multicolumn{2}{c|}{\textbf{ImageNet}} \\
        \vspace{-.06in} 
         &  \tiny ResNet & \tiny ResNet & \tiny WideResNet & \tiny DenseNet & \tiny ResNet & \tiny ResNet & \tiny DenseNet & \tiny WideResNet & \tiny ResNet & \tiny DenseNet \\
         &  \tiny 110 & \tiny 110\_SD & \tiny 32 & \tiny 40 & \tiny 110 & \tiny 110\_SD & \tiny 40 & \tiny 32 & \tiny 152 & \tiny 161 \\
     \hline
    \ecebin{} & \n& \y & \n & \y & \n & \y & \n & \n  & \n & \n  \\
    \cesweep{} & \y & \y & \y &\y  & \n & \y &\y & \y  & \n & \n   \\
    \bottomrule
    \end{tabular}
    \caption{The selection of a recalibration method is severely affected by biased computational methods.
    For ten models, we report whether either $\ecebin{}$ or $\cesweep{}$ select the same (\y) or different (\n) recalibration algorithm (either histogram binning, isotonic regression or temperature scaling) as would an estimate of TCE obtained from maximum likelihood fits to empirical data (see Section \ref{sec:glm_beta_fits}). $\ecebin{}$ selects the same algorithm only 3/10 times, versus 7/10 for $\cesweep{}$, illustrating how computational bias can negatively impact recalibration.
    }
    \label{tab:recalibration_selection}
\end{table*}

We address this problem by developing a technique to measure bias in calibration metrics, which we call the \emph{bias-by-construction (BBC)} framework.
The \BBC{} framework uses simulation to create a setting where the \tce{} can be computed analytically 
and thus the bias can be estimated directly.  \BBC{} reveals that
\ecebin{} has systematic non-negligible statistical bias, particularly for perfectly calibrated models
(Figure \ref{fig:intro}).

Our goal is to identify the least biased estimator of calibration error using \BBC{}.
We consider two estimators previously proposed in the literature: the \emph{debiased} estimator 
\citep{brocker2012,ferro2012}, which we refer to as \emph{\ecedebias}, and the smoothed kernel density estimator 
of \citet{zhang2020mix}, which we refer to as \emph{\kde}. Additionally, we propose an extension of 
\ecebin{} where the number of bins is chosen to ensure monotonicity of the calibration histogram,
which we refer to as \emph{\cesweep{}}.
\ecebin{}, \ecedebias{}, and \cesweep{} all require the binning of model confidence scores, and under
the lens of bias, we examine two common methods for specifying bins: partitioning the confidence-score
continuum either into \emph{equal width} bins or bins of \emph{equal mass}---equal numbers of data instances.


Furthermore, \BBC{} allows us to examine the impact of biased estimators in downstream decision making, such as the selection of a post-hoc recalibration method.
For example, when the choices for recalibration include histogram binning \citep{zadrozny2001obtaining}, temperature 
scaling \citep{guo2017calibration}, and isotonic regression \citep{zadrozny2002transforming},
Table~\ref{tab:recalibration_selection} illustrates that our bias-reduced measure, \cesweep{}, more frequently selects the `optimal' recalibration method when compared to the standard measure, \ecebin{} (70\% versus 30\% correctness, respectively). Optimality is determined by estimating TCE using numerical integration on curves arising from maximum likelihood fits across multiple model families, where we select the best model via the Akaike information criterion (see Section \ref{sec:glm_beta_fits}).


To summarize the contributions of this work, 
the core contribution is a simulation framework, bias by construction or \BBC{}, that allows us to identify
and characterize systematic bias in calibration error metrics for realistic models and data sets.
We show that estimation of calibration error by the predominant 
method, \ecebin{}, is biased, and paradoxically the bias is most severe for perfectly calibrated models.
Bias can lead not only to the mis-estimation of calibration error but also to the wrong choice of recalibration 
method, yielding a poorly calibrated model. Moreover, we find that the selection of hyperparameters 
for measuring calibration (e.g., number of bins) is under-appreciated and is absolutely critical. To 
address these issues, we propose \cesweep{}, a simple algorithm based on the monotonicity principle 
of calibration curves. 
We compare the bias of various estimators using predictions from four popular neural architectures 
and three data sets.  We find that \ecebin{} is more
biased than either \ecedebias{} or \cesweep{}, and of these two improved measures, \ecedebias{} performs 
better for perfectly calibrated models and \cesweep{} for miscalibrated models. 
Finally, our analyses provide rigorous empirical evidence that for all  binning-based estimators, equal-mass binning obtains a more accurate estimate of true calibration error. This finding gives strong guidance to revise the 
current practice of equal-width binning.

\section{RELATED WORK}
\label{sec:related_work}
\textbf{\ecebin{}.} \ecebin{} with 15 bins of equal width is currently the most popular way to measure calibration error in the literature \citep{naeini2015obtaining, guo2017calibration}.
An alternative but less popular implementation evaluates \ecebin{} using bins of equal mass, which partitions examples into bins that have an equal number of examples \cite{zadrozny2001obtaining,kumar2019verified}.
Recently, \citet{nixon2019measuring} observed that \ecebin{} with equal-mass-binning produces more stable rankings of recalibration algorithms, which is consistent with our conclusion that equal mass \ecebin{} is a less biased estimator of \tce{}.

\textbf{Sensitivity of 
\ecebin{} to implementation hyperparameters.}
Several works have pointed out that \ecebin{} is sensitive to implementation details. \citet{kumar2019verified} show that \ecebin{} increases with number of bins. \citet{nixon2019measuring} find that \ecebin{} is sensitive to several hyperparameters, including $\ell_p$ norm, number of bins, and binning technique. 
In contrast to prior work, we explicitly quantify estimation bias in simulation for realistic model outputs, and we show precisely how the \textit{bias} in \ecebin{} varies with the choice of sample size, model architecture, datasets, and implementation hyperparameters for \ecebin{} such as number of bins and binning method.

\textbf{Less biased metrics for calibration error.} Motivated by the sensitivity of \ecebin{} to implementation hyperparameters, recent work has proposed less biased estimates of TCE.  In particular,
\citet{ferro2012} and \citet{brocker2012}
propose a \textit{debiased estimator}, \ecedebias, which uses a jackknife technique to estimate the per-bin bias in the standard \ecebin{}, and subtracts off this bias to achieve a better binned estimate of the calibration error. Similarly, \citet{zhang2020mix} propose a smoothed Kernel Density Estimation (KDE) method for reducing bias when estimating calibration error. Relative to \cesweep{}, both \ecedebias{} and \kde{} have an additional hyperparameter (number of bins or kernel bandwidth, respectively).
We compare \cesweep{}, \ecedebias{}, and \kde, finding circumstances in which \cesweep{} and \ecedebias{} have relative advantages in bias reduction.

\textbf{Alternative definitions of calibration error.}
Researchers have studied alternatives notions of calibration error that are distinct from TCE (see Section \ref{sec:background} for a formal definition of TCE).
For example, \citet{widmann2019calibration} proposed a kernel-based calibration error, KCE, which has no explicit dependence on the model's calibration function. \citet{gupta2020calibration} propose a calibration error metric inspired by the Kolmogorov-Smirnov (KS) statistical test that estimates the maximum difference between cumulative probability distributions describing the model's confidence and accuracy.
The KS is similar to the maximum calibration error (MCE) \citep{naeini2015obtaining} in that it computes a worst-case
deviation between confidence and accuracy, but the KS is computed on the CDF, while the MCE uses binning and is computed on the PDF.  In contrast, TCE measures the \textit{average} difference between confidence and accuracy. 
Both the worst case and average difference are useful measures but may be applicable under different circumstances \citep{guo2017calibration}.

\textbf{Monotonicity in calibration curves.}
While \citet{zadrozny2002transforming} used calibration curve monotonicity to motivate isotonic regression for recalibration, they observed monotonic calibration curves empirically on only a handful of pre-deep learning models, and without theoretical justification. In contrast, our work is the first to suggest using monotonicity to improve \textit{calibration metrics}. We provide both theoretical and extensive empirical evidence that monotonic calibration curves arise in modern deep networks.

\section{BACKGROUND} 
\label{sec:background}
Consider a binary classification setup with input $X\in\mathcal{X}$, target output $Y = \{0,1\}$, and suppose we have a model $f: X \rightarrow [0,1]$ whose output represents a confidence score that the true label $Y$ is 1.

\textbf{True calibration error (\tce{}).} We define true calibration error as the $\ell_p$ norm difference between a model's predicted confidence and the true likelihood of being correct:\footnote{In our experiments, we measure calibration error using the $\ell_2$ norm because it increases the sensitivity of the error metric to extremely poorly calibrated predictions, which tend to be more harmful in applications.}

\vspace{-.3in}
\begin{equation}
\tce{}(f) = \left( 
\mathbb{E}_{X} \left[ \left\rvert  f(X) - \mathbb{E}_{Y}[Y|f(X)] \right \rvert  ^p \right]
\right)^\frac{1}{p} .
\label{theoretical_ce}
\end{equation}

\begin{figure}[t!]
  \newcommand{\ppwm}{0.3}
  \centering
    \includegraphics[width=\linewidth]{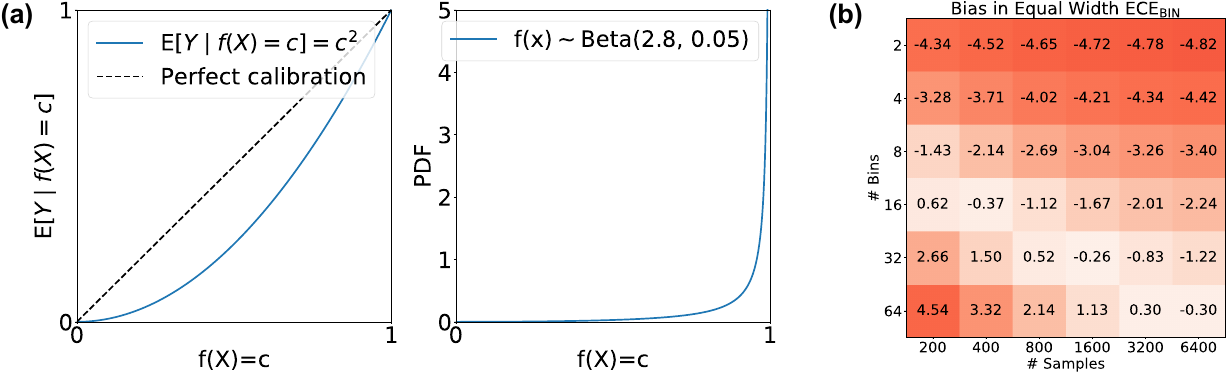}
\caption{(a) Curves controlling true calibration error.
 Our ability to measure calibration is contingent on both the confidence score distribution (e.g., $f(X)\sim\text{Beta}(2.8, 0.05)$) and the true calibration curve (e.g., $\truecurve = c^2$.
 (b) \ecebin{} may underestimate or overestimate \tce{}. The number of bins with minimal bias grows with the sample size.
  }
  \label{fig:tce_assumptions}
\end{figure}

Two independent features of a model determine \tce{}: (1) the distribution of confidence scores $f(x) \sim \mathcal{F}$ over which the outer expectation is computed, and (2) the true calibration curve \ensuremath{\E_Y[Y \mid f(X)]}, which governs the relationship between the confidence score $f(x)$ and the empirical accuracy (see Figure \ref{fig:tce_assumptions}a for illustration).


\subsection{Estimates of calibration error} To estimate the \tce{} of a model $f$, assume we are given a dataset containing $n$ samples, $\{x_i, y_i\}_{i=1}^n$. 
We can approximate \tce{} by replacing the outer expectation in Equation \ref{theoretical_ce} by the sample average and replacing the inner expectation with an average over instances with similar $f(x)$ values:

\vspace{-.3in}
\begin{equation}
\textstyle 
\ceneighborhood(f) = \left( 
\frac{1}{n} \sum_{i=1}^{n}
\left\rvert  f(x_i) - 
\frac{1}{|\mathcal{N}_i|} \sum_{j\in \mathcal{N}_i} y_j \right\rvert  ^p 
\right)^\frac{1}{p} ,
\label{ece_hat}
\end{equation}
where $\mathcal{N}_i$ is instance $i$'s set of neighbors in model confidence output space.

\textbf{Label-binned calibration error (\celb{}).} Label-binned calibration error uses binning to define $\mathcal{N}_i$ and estimate the model's empirical accuracy $\E[Y |f(X)]$.
The instances are partitioned
into $b$ bins, where $B_k$ denotes the
set of all instances in bin $k$, expressing
 Equation \ref{ece_hat} in terms of the binned neighborhood:
 \vspace{-.1in} 
\begin{equation}
\begin{aligned}
\textstyle 
\celb (f) &= \textstyle \left( 
\frac{1}{n} \textstyle \sum_{k=1}^{b}
\sum_{i\in B_k}
\left\rvert  f(x_{i}) - \bar{y}_k 
\right\rvert^p
\right)^\frac{1}{p} \hspace{-.1in}, \\
& \text{where }\bar{y}_k=\textstyle \frac{1}{|B_k|} \sum_{j\in B_k} y_j .
\label{ece_bn}
\end{aligned}
\end{equation}


\textbf{Binned calibration error (\ecebin{}).} In contrast to \celb{}, which operates on the original instances but uses binning to estimate empirical accuracy, \ecebin{} collapses all instances in a bin together and compares the per-bin empirical accuracy to the per-bin confidence score, weighted by the per-bin instance count.
Given $b$ bins, where $B_k$ is the set of instances in bin $k$, and letting $\bar{f}_k$ and $\bar{y}_k$ be the per-bin average confidence score and label, \ecebin{} is defined under the $\ell_p$ norm: 
\begin{equation}
\textstyle 
\ecebin{}(f) = \left( \sum_{k=1}^{b} \frac{|B_k|}{n}
\left\rvert  \bar{f}_k - \bar{y}_k 
\right\rvert^p \right)^\frac{1}{p}
\label{ece}
\end{equation}
Importantly, 
$\celb(f) \geq \ecebin(f)$,
which follows by applying Jensen's inequality on each inner term
$k \in \{ 1, 2, \ldots, b \}$
in Eqs.\ \ref{ece_bn} and \ref{ece}:
\begin{equation}
\textstyle 
\frac{1}{|B_k|}\sum_{i \in B_k} |f(X_i) - \bar{Y}_k|^p \ge
\left\rvert \sum_{i \in B_k} \bar{f}_k  - \bar{Y}_k
\right\rvert^p .
\label{ece_vs_celb}
\end{equation}

\section{THE BBC FRAMEWORK} 
\label{sec:simulation_setup}
We focus on bias rather than variance because the variance can be estimated from a finite set of samples through resampling techniques whereas the bias is an unknown quantity that reflects  systematic error. For completeness, we report variance for various calibration metrics as we vary the sample size, number of bins, and binning technique in Appendix \ref{apx:bias_and_variance}. We find empirically that the variance is relatively insensitive to the estimation technique and number of bins. 
 
The \textit{bias} of a calibration error estimator, $\text{ECE}_{\mathcal{A}}$ for some estimation algorithm $\mathcal{A}$, is the difference between the estimator's expected value with respect to the data distribution and the TCE:
\vspace{-.05in}
\begin{equation}
    \text{Bias}_\mathcal{A} = \E[\text{ECE}_{\mathcal{A}}] - \tce{}.
\label{eqn:true_bias}
\end{equation}
If we assume a specific confidence score distribution $\mathcal{F}$ and true calibration curve \mbox{$T(X)=\truecurve{}$} (see Figure \ref{fig:tce_assumptions}a for examples), we can compute the \tce{} by analytically or numerically evaluating the integral implicit in the outer expected value of Equation \ref{theoretical_ce}. 
We then compute a sample estimate of the bias by generating $n$ samples $\{f(x_i), y_i\}_{i=1}^n$ such that $f(x_i) \sim \mathcal{F}$ and $\truecurve{}:= T(c)$, and computing the ECE on the sample. We repeat this process for $m$ simulated datasets and compute the sample estimate of bias (hereafter, simply the ``bias'') as the difference between the average ECE and the TCE:
\vspace{-.02in}
 \begin{equation}
 \textstyle 
    \widehat{\text{Bias}}_\mathcal{A}(n) = \frac{1}{m} \sum_{i=1}^m \text{ECE}_{\mathcal{A}} - \tce{}.
\label{eqn:bias}
\vspace{-.10in}
\end{equation}

\label{sec:ece_bias_num_bins}


Using this \emph{bias-by-construction (\BBC)} framework, we next
investigate the bias in \ecebin{} as a function of the number of samples $n$ and the number of bins. We compute \ecebin{} with equal width binning and we assume parametric curves for $f(x)$ and $\truecurvex$ that are fit to the ResNet-110 CIFAR-10 model output. (Section \ref{sec:glm_beta_fits} has details on how we compute fits.)

Proposition 3.3 of \citet{kumar2019verified} asserts that any binned version of calibration error systematically underestimates \tce{} \textit{in the limit of infinite data}. However, for a finite number of samples $n$, Figure \ref{fig:tce_assumptions}b shows that \ecebin{} can either overestimate or underestimate \tce{} and that increasing the number of bins does not always lead to better estimates of \tce{}.
In Appendix \ref{apx:bias_and_variance}, we show how bias and variance vary for several calibration metrics as we change the binning scheme, sample size, and number of bins.
Regardless of binning scheme, for \ecebin{} we find empirically that \textit{there exists a bin number for each sample size that results in the lowest estimation bias and this optimal bin count grows with the sample size}. Intuitively, having a large number of bins is generally preferred because we can obtain a finer-resolution estimate of the calibration curve. However, if we have a small number of samples, setting the number of bins too high may result in a poor estimate of the calibration curve due to the low number of samples in each bin.

\section{MONOTONIC CALIBRATION METRICS}
\label{sec:monotonic_binning}
Though Section \ref{sec:ece_bias_num_bins} shows that there exists an optimal number of bins for which \ecebin{} has the lowest bias, unfortunately, this number depends on the binning technique, the number of samples, the confidence score distribution, and the true calibration curve. This observation motivates us to seek a method for adaptively choosing the number of bins.


Monotonicity in the true calibration curve implies that a model's expected accuracy should not decrease as the model's confidence increases. Although this
requirement seems reasonable for any statistical model, it is not obvious how to prove why or when a ``reasonable'' model would attain such a property. We offer a rationale for why it should be expected of machine learning models trained with a maximum likelihood objective, e.g., cross-entropy or logistic loss. Namely, from ROC (receiver operating characteristic) analysis of maximum likelihood models, an under-appreciated observation of ROC curves is that a model trained to maximize the likelihood ratio must have a convex ROC curve in the limit of infinite data \citep[see][Sec. 2.3]{green1966signal}. The slope of the ROC curve is related to the calibration curve, and a convex ROC curve implies a monotonically increasing calibration curve (the converse is also true) \citep{chen2018calibration, gneiting2018receiver}.

In practice, several potential confounds may lead to observing non-monontonic calibration curves. First, finite data size may lead to fluctuations in the true positive or false positive rates, but do not reflect the behavior of the underlying model. Second, deviations in domain statistics between cross-validated splits in the data may lead to unbounded behavior; however, we assume that such domain shifts are negligible as cross-validated splits are presumed to be selected {\it i.i.d.}.\footnote{Note that a third potential reason for a non-monotonic calibration curve is that a classifier could be trained with a non--likelihood-based statistical criteria, e.g. moment matching. However, a lack of monotonic behavior in the calibration curve of such a model may actually be a sign that the model is not reasonable or admissible model on a given task \citep{chen2018calibration, pesce2010convexity}.} Given that deviations from non-monotonic calibration curves are considered artificial, we posit that any method that is trying to assess the TCE of an underlying model may freely assume monotonicity in the true calibration curve.
Note that this proposition already guides the entire field of re-calibration to require that re-calibration methods only consider monotonic functions \citep{platt1999probabilistic, zadrozny2002transforming, wu2012spline}.


Accordingly, we leverage underlying monotonicity in the true calibration curve and propose the \emph{monotonic sweep calibration error}, a metric that chooses the largest number of bins possible such that the chosen bin size and all smaller bin sizes preserve monotonicity in the bin heights $\bar{y}_k$, i.e., 
\begin{equation}
\begin{aligned}
\textstyle 
\cesweep{} =  \left( \sum_{k=1}^{b^*} \frac{|B_k|}{n} \left\rvert  \bar{f}_k - \bar{y}_k \right\rvert^p \right)^\frac{1}{p}  \text{~where~} \\
b^* = \max \{ b ~|~ 1 \le b \le n ; \forall~ b' \le b,  ~\bar{y}_1 \le \ldots \le \bar{y}_{b'} \}
\end{aligned}
\label{ece_seep}
\vspace{-10pt}
\end{equation}

\begin{algorithm}[ht!]
\vspace{-3pt}
\setlength{\intextsep}{1\baselineskip}
\caption{Monotonic Sweep Calibration Error}
\label{alg:mon_sweep}
\begin{algorithmic}
\FOR{$b=2$ {\bfseries to} $n$}
\STATE Compute bin heights ($\bar{y}_k$) for \ecebin{} using $b$  bins
\IF{binning is not monotonic}
\STATE $b \leftarrow b - 1$
\STATE \textbf{break}
\ENDIF
\ENDFOR
\STATE \textbf{return} \ecebin{} computed with $b$ bins
\end{algorithmic}
\vspace{-3pt}
\end{algorithm}

We compute the monotonic sweep calibration error by starting with $b=2$ bins ($b=1$ is guaranteed to be a monotonic binning) and gradually increasing the number of bins until we either reach a non-monotonic binning, in which case we return the last $b$ that corresponded to a monotonic binning, or until every sample belongs to its own bin ($b=n$).  In Appendix \ref{apx:chosen_bins}, we explore the number of bins chosen by \cesweep{} for varying sample sizes and model output.

\begin{figure*}[t]
  \centering
  \includegraphics[width=.8\linewidth]{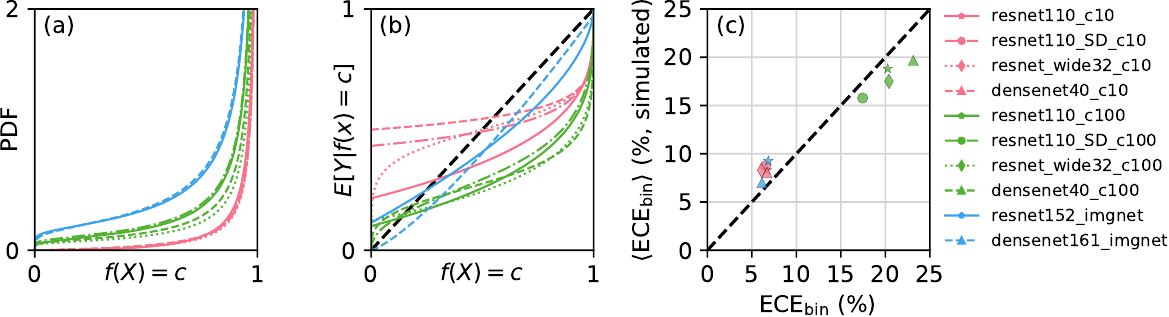}
  
  \caption{ \label{fig:beta_glm_fit} {Maximum likelihood fits to empirical datasets illustrate large skew in their density distribution and calibration function}. For each dataset, we fit (a) confidence distributions with Beta distribution and (b) calibration curves with generalized linear models across multiple model families, selecting the best model via the Akaike information criterion (details in Appendix~\ref{apx:glm_beta_modeling_appendix}). We find the dataset source has a greater influence over the curves than the model architecture. (c) We plot the overall quality of the fits by computing the \ecebin{} on the original data vs. the \ecebin{} averaged over $1000$ simulated trials. Curves well-fit to the data lie close to the identity line.}
\end{figure*}

\section{FITTING THE CALIBRATION CURVE AND SCORE DISTRIBUTION}
\label{sec:glm_beta_fits}  
TCE is analytically computable when we assume parametric forms for the confidence distribution and the true calibration curve. In order to ensure that the parametric forms we use in simulation reflect the diversity and complexity of realistic model output, we develop parametric models of empirical logit datasets.

We consider 10 publicly available logit datasets \citep{nn_markus} that arise from training four different architectures
(ResNet, ResNet-SD, Wide-ResNet, and DenseNet) \citep{he2016deep,huang2016deep,zagoruyko2016wide,huang2017densely,lecun1998gradient} on three different image datasets (CIFAR-10/100 and ImageNet) \citep{Krizhevsky2009LearningML, imagenet_cvpr09}.
For each example in a given dataset, we compute top-label confidence scores by selecting the maximum softmax score across all classes and we compute whether or not the example resulted in a ``hit,'' i.e. whether the model's predicted class corresponds to the true class.  By using only the top-label confidence score and determining whether the top and true labels match, we can treat the calibration problem as binary.

For the parametric fits, we model confidence score distributions $f(X)$ using a beta density fit via maximum likelihood estimation. The beta distribution is a flexible continuous probability distributions on the interval [0,1], which makes it a natural choice for representing the probability distribution defined by the model output. For calibration curves, we fit multiple (binary) generalized linear models (GLM) to the top-label output and then select the best model using the Akaike Information Criteria (AIC). The AIC is a standard procedure for model selection in the literature, for selecting the model that most adequately describes data arising from a mechanism included in the model family.
The GLM models considered include logit, log, and "logflip" ($\text{log}(1-x)$) link and transformation functions, up to first order in the transformed domain, which all result in monotonic calibration functions. 
See Appendix~\ref{apx:glm_beta_modeling_appendix} for additional details.

We find that the parametric forms for the calibration curve and distribution of scores are well captured by simple GLM and beta models. Figures~\ref{fig:beta_glm_fit}a,b show the resulting fits with parameters summarized in Appendix \ref{apx:glm_beta_modeling_appendix}. 
We observe significant skew in the score distribution which, as discussed in Section~\ref{sec:simulation_glm_beta}, poses a challenge to measuring calibration error with equal-width bins.  We find that the dataset has more influence on the fits than the neural model, with ImageNet models the least skewed and CIFAR-10 the most (correlating with model accuracy).
Figure~\ref{fig:beta_glm_fit}c demonstrates that \ecebin{} scores computed on simulated data from the fits closely match \ecebin{} scores computed on the real data.

\section{RESULTS}
\label{sec:main}
\subsection{Estimating bias on real models and data}
\label{sec:simulation_glm_beta}
\begin{figure*}[t]
  \centering
  \includegraphics[width=.8\linewidth]{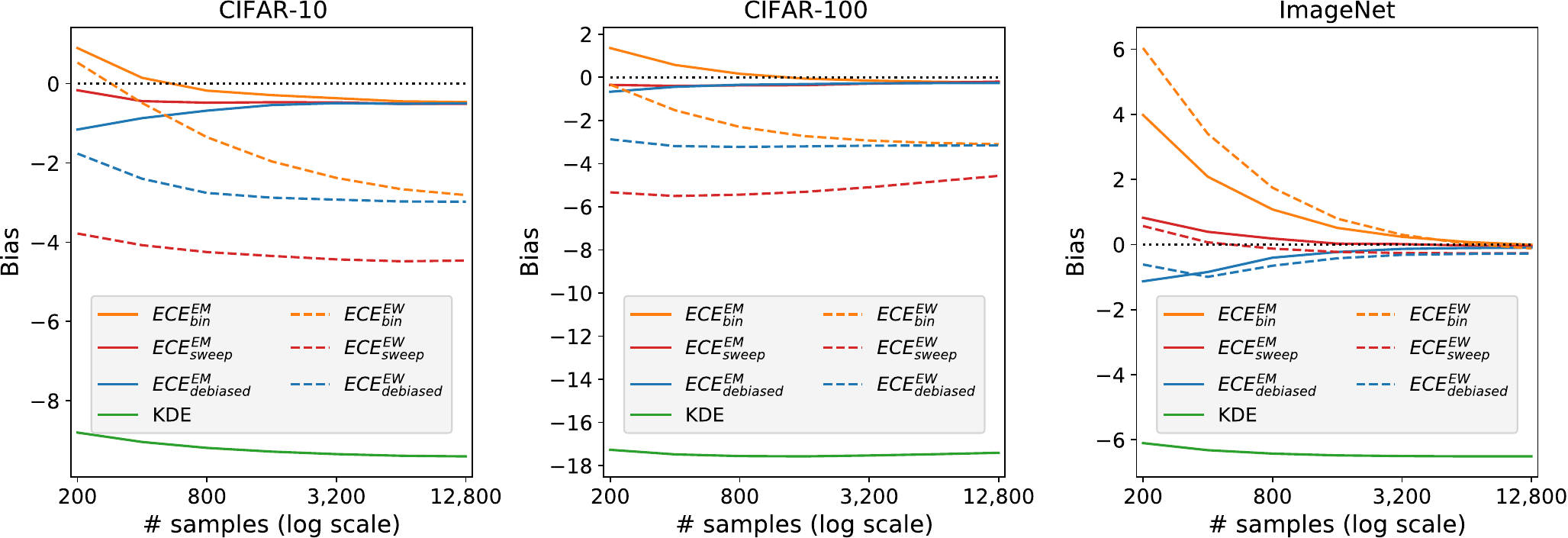}
  \caption{{\emmonsweep{} is less biased than alternative calibration metrics.}
  We plot bias versus number of samples $n$ for calibration metrics on simulated data drawn from the CIFAR-10, CIFAR-100, and ImageNet fits (Section \ref{sec:glm_beta_fits}). The dataset the model was trained on has a greater influence on bias than the model architecture. Metrics based on equal mass binning consistently outperform equal width binning. Exploiting monotonicity in the \emmonsweep{} metric helps the most at small sample sizes.}
  \label{fig:ml_bias}
\end{figure*}

Our bias-by-construction (\BBC) framework uses the parametric fits to real models
and datasets from Section \ref{sec:glm_beta_fits} to estimate bias as follows.
Each fit permits the analytical or numerical computation of
TCE and can also be used in generative fashion to draw a synthetic
set of examples. ECE can then be estimated from these samples, and the 
difference between the estimated ECE and TCE across many samples---$1{,}000$ in
results to be presented---yields the bias (Equation~\ref{eqn:bias}).

We estimate bias for \ecebin{}, \ecedebias{}, and \cesweep{} using both equal-mass 
and equal-width binning, and also for the \kde{} 
estimator.  Following \citet{guo2017calibration}, we choose 15 bins for 
\ecebin{} and \ecedebias{}. (Appendix \ref{apx:bias_and_variance} includes
an analysis that varies the number of bins and finds that the optimal number 
of bins for bias minimization depends on the number of samples. This 
Appendix also includes calculations of variance across estimators, bin numbers, and 
sample sizes.)

Figure \ref{fig:ml_bias} plots the bias
versus sample size for seven estimators, shown separately
for each of three datasets. Because the curves for
individual architectures look very similar to one another for a given data
set, we have averaged over model architectures. The black dotted line 
indicates an unbiased estimator. 

\textbf{Equal-width versus equal-mass binning.} 
The dashed and solid lines correspond to equal width (\EW) and equal mass (\EM) 
bins, respectively, and the colors indicate the metric. For the three 
binning-based metrics, \EM{} consistently obtains a smaller magnitude bias than \EW.
This finding is not well appreciated in the literature: \EW{} is the common practice.
For instance, \citet{kumar2019verified} proposed \ewdebias{} and did not consider \emdebias{}. However, our results show that \emdebias{} is a consistently less biased estimator than \ewdebias{}.
Our work therefore provides immediate and strong guidance to researchers and
practitioners concerned with model calibration. An explanation
for the advantage of \EM{} over \EW{} stems from the fact that, as shown in 
Figure \ref{fig:beta_glm_fit}a,  models trained on CIFAR-10 and CIFAR-100 have highly skewed confidence distributions. 
Consequently, \EW{} binning places most instances in the top bin. 
As we increase the number of samples, we increase the likelihood that we 
generate a sample that populates one of the lower bins, which, due to their 
low sample density, may have a poorer average estimate of the \tce{}. On ImageNet, where the confidence distribution is less skewed, the advantage of \EM{} over \EW{} is still consistent but less pronounced.

\textbf{Comparing metrics.} Across the three datasets and various sample sizes,
\emmonsweep{} appears to perform the best.  \emdebias{} also performs well but
not as well as \emmonsweep{} at low sample sizes.
To determine whether the difference between \emmonsweep{} and \emdebias{} is statistically reliable, we conducted a paired $t$-test on absolute bias.
Across datasets, models, and number of samples, we find a statistically significant difference: a mean absolute bias is 0.504 for \emdebias{} and 0.347 for \emmonsweep{} (t = 5.10, p < $1\mathrm{e}{-5}$). Appendix \ref{apx:bias_and_variance} demonstrates    higher variance for \emdebias{} than \emmonsweep{}.


The \kde{} metric has much larger bias across the three datasets than any of the other metrics. This finding suggests that the heuristic used to choose the kernel bandwidth and the specific `triweight’ kernel worked well for the synthetic example evaluated in \citet{zhang2020mix}, but fails to generalize to the more realistic examples we study. Specifically, \citet{zhang2020mix} assume a Gaussian distribution for $P(X | Y)$ and a logistic confidence score distribution, which result in notably different qualitative shapes than the logit distributions we obtain from models trained on CIFAR-10/100 or ImageNet (see Figure  \ref{fig:beta_glm_fit}a,b or the reliability diagrams and score distributions of \citeauthor{nn_markus}, \citeyear{nn_markus}).





\subsection{How well can we detect miscalibration?}
\label{sec:miscalibration}
Pragmatically, practitioners may be less concerned about bias per se than being able to answer a
straightforward question about a model: is the model miscalibrated?  If the validation set provides
clear evidence of miscalibration, further steps must be taken to correct the miscalibration. However,
given bias in the ECE metrics, the mere observation of an ECE > 0 is not sufficient to raise alarm.

\begin{SCfigure}[1][b!]
\centering
\includegraphics[height=1.7in]{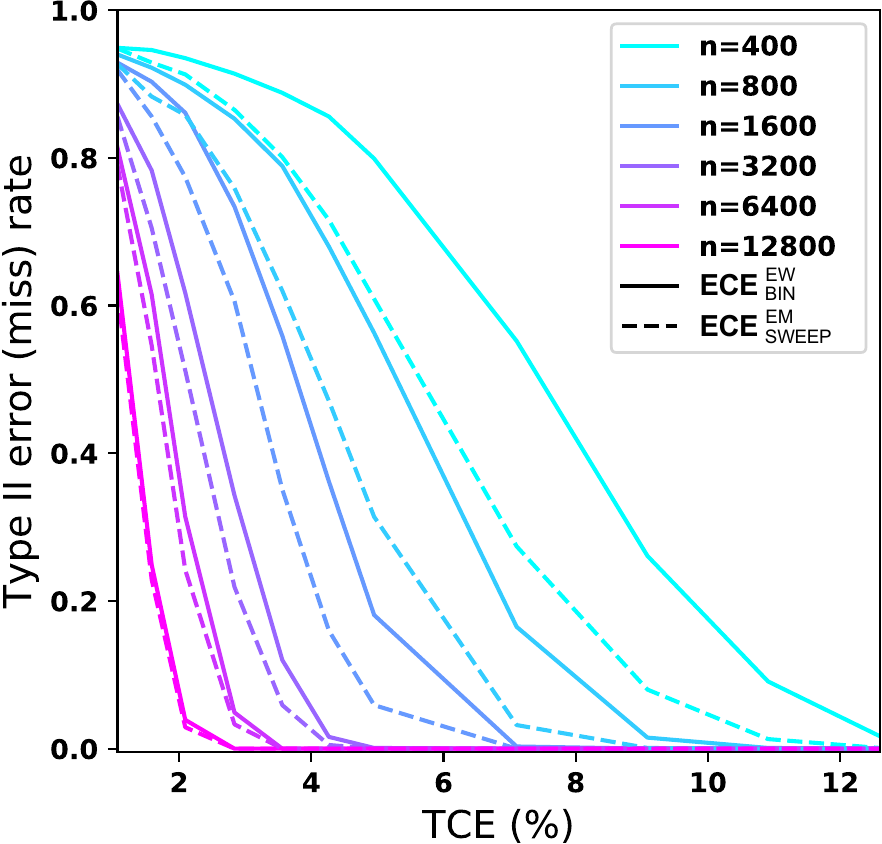}
\caption{Probability of failing to detect miscalibration (miss rate) as a function of TCE and sample size ($n$).
\emmonsweep{} has lower failure rate than \ewece{}.}
\label{fig:hyptest}
\vspace{-10pt}
\end{SCfigure}

\label{sec:bias_vs_true_calib_err}
\begin{figure*}[t]
  \centering
  \includegraphics[width=.92\linewidth]{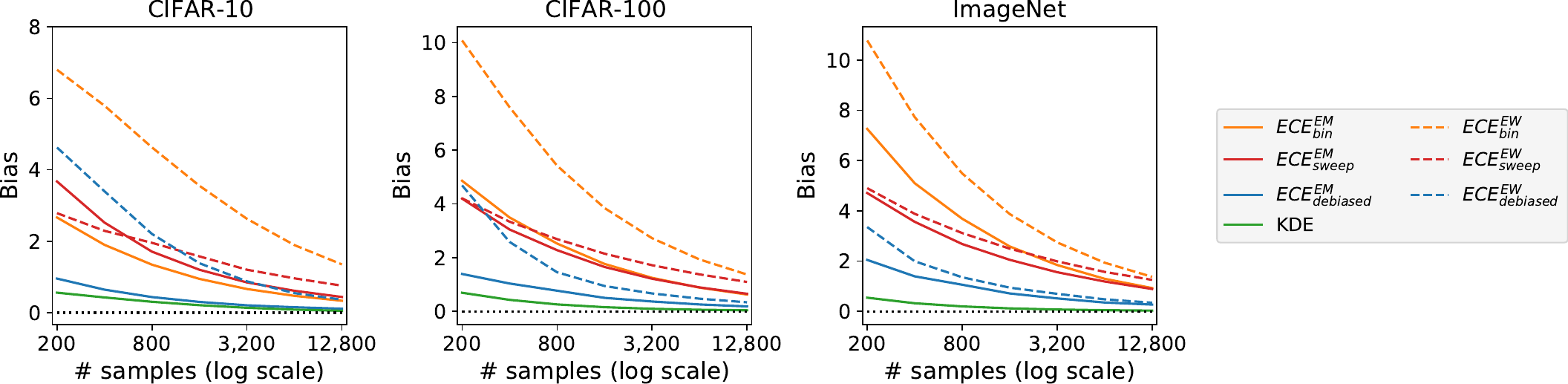}
  \caption{Perfectly calibrated models. Using \BBC{} we set the calibration curve to have 0\% calibration error but use realistic model confidence score distributions. We find that the KDE estimator has the least amount of bias for perfectly calibrated models, followed by \emdebias{}.}
  \label{fig:calib_compare}
\end{figure*}



Consider the situation with a model of unknown TCE, and we wish to perform hypothesis testing to determine if we can reject the null hypothesis that TCE=0. Our ability to detect miscalibration depends on TCE, the sample size ($n$),
and the method for estimating calibration error. We conduct a simulation with $f(x) \sim \text{Beta}(1,1)$ and true
calibration curve from the family $\truecurve{}=c^d$, where $d$ is varied to obtain a range of TCE. Allowing for
a type I error rate of .05 (also known as the false-alarm rate, or the rate of mistakenly claiming miscalibration despite perfect calibration), we obtain type II error rates (also known as the miss rate, or the rate of failing to detect
a miscalibration). Figure~\ref{fig:hyptest} shows the type II error rate as a function of TCE and $n$ for 
the metric typically used in practice (\ewece{}) and the best performing metric identified in the previous section (\emmonsweep{}).  \emmonsweep{} obtains a significantly lower failure rate
than \ewece{}, particularly for under $10{,}000$ samples. More generally, we note limitations with both methods: to detect
a miscalibration of 2\%, over $10{,}000$ samples are needed; and if one has under 500 samples, the miscalibration must be greater 
than 10\% to be detected reliably.

\subsection{Perfect calibration}

In Section \ref{sec:simulation_glm_beta}, we studied realistic scenarios of models whose outputs have the
same statistics as common neural architectures on popular datasets. The BBC framework also allows us to
simulate a continuum of models that differ systematically in TCE. For all metrics, bias increases
as TCE decreases (details in Appendix~\ref{apx:controlling_tce}). This finding is not surprising because 
binned metrics always produce a nonnegative ECE estimate, and in the limit of a perfectly calibrated model,
any deviation of the binning histogram from the diagonal will result in positive bias.

In this section, we compare the bias of estimators for the case of a perfectly calibrated model---the 
ultimate aim of designing methods that minimize miscalibration.  To simulate perfect calibration,
the calibration curve of the model is set to $\truecurve = c$, but we use the realistic confidence 
score distributions from the previous section. 

Figure \ref{fig:calib_compare} illustrates the effect of sample size on bias for the seven different 
estimators under perfect calibration. Although the KDE estimator outperforms all others, it is not a 
viable candidate because it has a very high bias for realistic scenarios (Figure~\ref{fig:ml_bias}).
Excluding KDE, \emdebias{} is the least biased metric, obtaining significantly lower bias than \emmonsweep{}.

How do we reconcile these results with our previous finding (Figure~\ref{fig:ml_bias}) that \emmonsweep{}
is preferred over \emdebias{}? The present results assume a well calibrated model; the previous results
are based on realistic scenarios. Whether one prefers \emmonsweep{}
or \emdebias{} ultimately depends on a practitioner's prior beliefs about a model's degree of miscalibration.
But to some degree we are splitting hairs: both \emmonsweep{} and \emdebias{} are consistently superior to
common practice (\ewece{}) and proposed improvements (e.g., \ewdebias{}, as recommended by
\citeauthor{kumar2019verified}, \citeyear{kumar2019verified}).

%
%

\section{DISCUSSION AND CONCLUSION}
\label{sec:discussion}
Calibration research typically focuses on recalibrating models, i.e., transforming $f(x)$ to $f'(x)$ \citep{platt1999probabilistic, zadrozny2001obtaining, zadrozny2002transforming}.  We focus on estimating true calibration error, because without a good estimate, how is one to select and evaluate recalibration methods? The preferred recalibration method for a given model and data set is affected by bias: Table~\ref{tab:recalibration_selection} shows that using \emmonsweep{} to select a recalibration method instead of \ewece{} leads to better choices and subsequently, better calibration on the test set. Indeed,
bias may have impacted the conclusions of previous studies of calibration error, such as the well cited work
of \citet{guo2017calibration}.
The choice of calibration error estimator can also impact the detection of miscalibration: Figure~\ref{fig:hyptest} indicates that \emmonsweep{} is a more sensitive metric than \ewece{} for detecting if a model is miscalibrated.

 Several authors attempt a different approach to recalibration: improving model calibration during training. For instance, \citet{mukhoti2020calibrating} train a model with a batch size of 128 across multiple types of losses including maximum mean calibration error \citep{kumar2018trainable} and Brier loss \citep{brier1950verification} which explicitly minimizes calibration loss using 128 examples at a time.
 However, our results suggest that training a model with naive estimates of calibration error using a batch size < $O(1000)$ is a potentially flawed endeavor, particularly because the distribution of scores from the model changes throughout training, and any potential calibration measure may be more affected by the distribution of scores than the true calibration curve.
  
Our work can be extended in many directions which we did not have space here to consider, including: examining violations of our distributions assumptions and the setting where the confidence-score distributions are less skewed; studying the interplay between bias and variance; exploring alternative task types, such as binary classification; and evaluating alternative calibration measures such as $\ell_1$ TCE and KCE, KS, and MCE.

Relying on the predictions from machine learning models in high stakes situations like autonomous vehicles, content moderation, and medicine, requires the ability to detect predictions that are likely to be incorrect. Given that the default confidence scores produced by machine learning models do not necessarily correspond to the model’s empirical accuracy, recalibration is necessary in order to produce reliable and consistent output.  However, it is impossible to perfectly calibrate a model if calibration cannot be measured accurately. Our results show that the statistical bias in current calibration error estimators grows as we approach perfect calibration,  but this bias can be mitigated by using equal-mass binning and methods such as the debiased estimator \citep{brocker2012,ferro2012,kumar2019verified}, \emdebias{}, or our own monotonic estimation technique, \emmonsweep{}.




\section*{Acknowledgements}
We would like to thank Simon Kornblith, Jize Zhang, Tengyu Ma, and Ananya Kumar for helpful comments on this work.

\nocite{gelmanbda04} 

\bibliography{references}
\bibliographystyle{apalike} 

\newpage
\clearpage
\appendix

\section{Maximum-likelihood fits}
\label{apx:glm_beta_modeling_appendix}

\subsection{Confidence score distribution fits}

Table \ref{table:beta_fit_data} provides parameters of best fit for the Beta distribution for each of 10 empirical datasets, obtained by fitting the top-label confidence score via maximum likelihood estimation. 
\begin{table}[ht!]
\centering
\caption{Parameters of best fit for Beta distribution investigated in Section \ref{sec:glm_beta_fits}.}
\label{table:beta_fit_data}
\begin{tabular}{lrr}
\toprule
{} &  $\hat{\alpha}$ &  $\hat{\beta}$ \\
\midrule
resnet110\_c10      &     2.7752 &    0.0478 \\
resnet110\_SD\_c10   &     2.1714 &    0.0394 \\
resnet\_wide32\_c10  &     2.3806 &    0.0379 \\
densenet40\_c10     &     1.9824 &    0.0397 \\
resnet110\_c100     &     1.1823 &    0.1081 \\
resnet110\_SD\_c100  &     1.1233 &    0.1147 \\
resnet\_wide32\_c100 &     1.0611 &    0.0650 \\
densenet40\_c100    &     1.0805 &    0.0808 \\
resnet152\_imgnet   &     1.1359 &    0.2069 \\
densenet161\_imgnet &     1.1928 &    0.2206 \\
\bottomrule
\end{tabular}
\end{table}

Global optimia $\hat{\alpha} \in [0,200]$, $\hat{\beta} \in [0,50]$ are approximately computed using a recursively-refining brute-force search until both parameters are established to within an absolute tolerance of $1\mathrm{e}{-5}$. 
Each step in the recursion contracts a linear sampling grid ($N=11$) by a factor of $\gamma=.5$ centered on the previously established optimal parameter, subject to the constraints $\alpha, \beta>0$.
Experiments confirmed that the computed optima were robust to the hyperparameters $N, \gamma$.

\begin{equation}
  \argmin_{\alpha, \beta} \sum_i -\ln \frac{  x_{i}^{\alpha-1}  (1-x_{i})^{\beta-1} }{ \frac{\Gamma\left({\alpha}\right) \Gamma\left({\beta}\right)}{\Gamma\left(\alpha+\beta \right)} }
  \label{eq:beta_dist_mle_obj}
\end{equation}
 
\subsection{Calibration curve fits}
\begin{table*}
  \centering
  \caption{Parameters of best fit for calibrations functions investigated in Section \ref{sec:glm_beta_fits} (table continues on next page).}
  \label{table:glm_fit_data}
  \scalebox{0.8}{
  \begin{tabular}{ c }
  \begin{tabular}{llrrr}
\toprule
               &            &                         AIC &  $b_{0}$ &  $b_{1}$ \\
dataset\_name & glm\_name &                             &         &         \\
\midrule
resnet110\_c10 & logflip\_logflip\_b0\_b1 & \textbf{2779.22} &   -0.24 &    0.30 \\
               & logit\_logflip\_b0\_b1 &                     2790.40 &   -0.55 &   -0.38 \\
               & logit\_logflip\_b1 &                     2827.51 &      &   -0.31 \\
               & logit\_logit\_b0\_b1 &                     2840.70 &   -0.38 &    0.36 \\
               & logit\_logit\_b1 &                     2900.02 &      &    0.30 \\
               & logflip\_logflip\_b1 &                     2932.09 &      &    0.34 \\
               & log\_log\_b0\_b1 &                     3221.72 &   -0.06 &    2.53 \\
               & logit\_logit\_b0 &                     3799.63 &    1.99 &      \\
               & logflip\_logflip\_b0 &                     3811.98 &   -2.13 &      \\
               & log\_log\_b0 &                     3829.05 &   -0.13 &      \\
               & logit\_logflip\_b0 &                     3868.40 &    1.95 &      \\
               & log\_log\_b1 &                     4281.78 &      &    4.75 \\
resnet110\_SD\_c10 & logit\_logflip\_b0\_b1 & \textbf{2498.98} &   -0.27 &   -0.35 \\
               & logit\_logit\_b1 &                     2502.52 &      &    0.30 \\
               & logit\_logflip\_b1 &                     2508.70 &      &   -0.30 \\
               & logit\_logit\_b0\_b1 &                     2538.41 &   -0.26 &    0.33 \\
               & logflip\_logflip\_b0\_b1 &                     2550.29 &   -0.36 &    0.27 \\
               & logflip\_logflip\_b1 &                     2572.85 &      &    0.35 \\
               & log\_log\_b0\_b1 &                     2594.91 &   -0.08 &    1.98 \\
               & log\_log\_b0 &                     3137.19 &   -0.19 &      \\
               & logflip\_logflip\_b0 &                     3150.42 &   -1.80 &      \\
               & logit\_logit\_b0 &                     3175.58 &    1.58 &      \\
               & logit\_logflip\_b0 &                     3179.67 &    1.56 &      \\
               & log\_log\_b1 &                     3697.37 &      &    3.77 \\
resnet\_wide32\_c10 & logit\_logit\_b1 & \textbf{2483.34} &      &    0.26 \\
               & logflip\_logflip\_b0\_b1 &                     2487.69 &   -0.47 &    0.22 \\
               & logit\_logit\_b0\_b1 &                     2511.39 &   -0.13 &    0.28 \\
               & logit\_logflip\_b0\_b1 &                     2558.45 &   -0.26 &   -0.28 \\
               & logit\_logflip\_b1 &                     2586.47 &      &   -0.25 \\
               & log\_log\_b0\_b1 &                     2647.03 &   -0.12 &    1.87 \\
               & logflip\_logflip\_b1 &                     2713.17 &      &    0.30 \\
               & log\_log\_b0 &                     2981.24 &   -0.21 &      \\
               & logflip\_logflip\_b0 &                     2983.05 &   -1.70 &      \\
               & logit\_logit\_b0 &                     2989.90 &    1.49 &      \\
               & logit\_logflip\_b0 &                     3055.55 &    1.45 &      \\
               & log\_log\_b1 &                     4582.09 &      &    4.61 \\
densenet40\_c10 & logit\_logflip\_b1 & \textbf{2910.62} &      &   -0.26 \\
               & logit\_logit\_b0\_b1 &                     2961.31 &   -0.40 &    0.31 \\
               & logit\_logflip\_b0\_b1 &                     3000.23 &   -0.38 &   -0.31 \\
               & logflip\_logflip\_b0\_b1 &                     3001.78 &   -0.31 &    0.24 \\
               & logit\_logit\_b1 &                     3021.54 &      &    0.25 \\
               & logflip\_logflip\_b1 &                     3027.78 &      &    0.31 \\
               & log\_log\_b0\_b1 &                     3153.38 &   -0.12 &    2.04 \\
               & log\_log\_b0 &                     3531.22 &   -0.22 &      \\
               & logflip\_logflip\_b0 &                     3589.11 &   -1.60 &      \\
               & logit\_logit\_b0 &                     3601.85 &    1.37 &      \\
               & logit\_logflip\_b0 &                     3679.95 &    1.30 &      \\
               & log\_log\_b1 &                     4735.18 &      &    4.27 \\
resnet110\_c100 & logflip\_logflip\_b0\_b1 & \textbf{8181.97} &   -0.11 &    0.28 \\
               & logit\_logit\_b0\_b1 &                     8206.19 &   -0.88 &    0.39 \\
               & logflip\_logflip\_b1 &                     8301.28 &      &    0.31 \\
               & logit\_logflip\_b0\_b1 &                     8371.53 &   -1.01 &   -0.40 \\
               & logit\_logit\_b1 &                     8732.11 &      &    0.25 \\
               & log\_log\_b0\_b1 &                     8918.21 &   -0.16 &    2.35 \\
               & logit\_logflip\_b1 &                     8926.99 &      &   -0.23 \\
               & logit\_logflip\_b0 &                    10903.83 &    0.74 &      \\
               & logit\_logit\_b0 &                    10943.95 &    0.72 &      \\
               & logflip\_logflip\_b0 &                    10964.91 &   -1.12 &      \\
               & log\_log\_b0 &                    11002.20 &   -0.40 &      \\
               & log\_log\_b1 &                    11850.89 &      &    4.26 \\
\bottomrule
\end{tabular}

  \end{tabular}
  }
\end{table*}

\begin{table*}
  \centering
  \scalebox{0.8}{
  \begin{tabular}{ c }
  \begin{tabular}{llrrr}
\toprule
                   &            &                          AIC &  $b_{0}$ &  $b_{1}$ \\
dataset\_name & glm\_name &                              &         &         \\
\midrule
resnet110\_SD\_c100 & logit\_logit\_b0\_b1 &  \textbf{7873.61} &   -0.88 &    0.49 \\
                   & logflip\_logflip\_b0\_b1 &                      7878.19 &   -0.09 &    0.35 \\
                   & logflip\_logflip\_b1 &                      7932.28 &      &    0.38 \\
                   & logit\_logflip\_b0\_b1 &                      7944.61 &   -1.04 &   -0.52 \\
                   & logit\_logit\_b1 &                      8315.51 &      &    0.32 \\
                   & log\_log\_b0\_b1 &                      8437.82 &   -0.11 &    2.18 \\
                   & logit\_logflip\_b1 &                      8510.36 &      &   -0.30 \\
                   & log\_log\_b1 &                      9988.07 &      &    3.30 \\
                   & logit\_logit\_b0 &                     10803.27 &    0.80 &      \\
                   & log\_log\_b0 &                     10810.90 &   -0.37 &      \\
                   & logflip\_logflip\_b0 &                     10823.15 &   -1.16 &      \\
                   & logit\_logflip\_b0 &                     10834.48 &    0.78 &      \\
resnet\_wide32\_c100 & logflip\_logflip\_b0\_b1 &  \textbf{7183.93} &   -0.13 &    0.21 \\
                   & logit\_logit\_b0\_b1 &                      7219.14 &   -0.98 &    0.33 \\
                   & logflip\_logflip\_b1 &                      7233.51 &      &    0.25 \\
                   & logit\_logflip\_b0\_b1 &                      7297.00 &   -1.06 &   -0.34 \\
                   & logit\_logit\_b1 &                      7626.21 &      &    0.19 \\
                   & log\_log\_b0\_b1 &                      7650.97 &   -0.24 &    2.51 \\
                   & logit\_logflip\_b1 &                      7795.28 &      &   -0.17 \\
                   & logflip\_logflip\_b0 &                      8977.39 &   -0.98 &      \\
                   & logit\_logflip\_b0 &                      8987.38 &    0.49 &      \\
                   & log\_log\_b0 &                      9000.24 &   -0.49 &      \\
                   & logit\_logit\_b0 &                      9009.51 &    0.49 &      \\
                   & log\_log\_b1 &                     11911.51 &      &    5.48 \\
densenet40\_c100 & logit\_logit\_b0\_b1 &  \textbf{8158.28} &   -0.97 &    0.34 \\
                   & logflip\_logflip\_b0\_b1 &                      8229.43 &   -0.12 &    0.22 \\
                   & logit\_logflip\_b0\_b1 &                      8267.77 &   -1.08 &   -0.35 \\
                   & logflip\_logflip\_b1 &                      8368.86 &      &    0.25 \\
                   & logit\_logit\_b1 &                      8783.50 &      &    0.19 \\
                   & log\_log\_b0\_b1 &                      8832.20 &   -0.25 &    2.26 \\
                   & logit\_logflip\_b1 &                      8918.57 &      &   -0.18 \\
                   & logit\_logit\_b0 &                     10138.24 &    0.47 &      \\
                   & logit\_logflip\_b0 &                     10182.61 &    0.45 &      \\
                   & logflip\_logflip\_b0 &                     10242.15 &   -0.94 &      \\
                   & log\_log\_b0 &                     10261.01 &   -0.50 &      \\
                   & log\_log\_b1 &                     13322.10 &      &    5.25 \\
resnet152\_imgnet & logflip\_logflip\_b0\_b1 & \textbf{18729.85} &   -0.12 &    0.58 \\
                   & logit\_logit\_b0\_b1 &                     18783.22 &   -0.29 &    0.65 \\
                   & log\_log\_b0\_b1 &                     18785.44 &   -0.03 &    1.32 \\
                   & logflip\_logflip\_b1 &                     18872.14 &      &    0.65 \\
                   & logit\_logit\_b1 &                     19074.37 &      &    0.57 \\
                   & logit\_logflip\_b0\_b1 &                     19095.40 &   -0.82 &   -0.79 \\
                   & log\_log\_b1 &                     19840.25 &      &    1.53 \\
                   & logit\_logflip\_b1 &                     20062.10 &      &   -0.50 \\
                   & logflip\_logflip\_b0 &                     26935.09 &   -1.41 &      \\
                   & log\_log\_b0 &                     26968.50 &   -0.28 &      \\
                   & logit\_logflip\_b0 &                     27012.77 &    1.12 &      \\
                   & logit\_logit\_b0 &                     27084.11 &    1.11 &      \\
densenet161\_imgnet & log\_log\_b0\_b1 & \textbf{18202.41} &   -0.03 &    1.27 \\
                   & logit\_logit\_b0\_b1 &                     18460.70 &   -0.25 &    0.68 \\
                   & logflip\_logflip\_b1 &                     18521.48 &      &    0.67 \\
                   & logflip\_logflip\_b0\_b1 &                     18534.07 &   -0.10 &    0.61 \\
                   & logit\_logit\_b1 &                     18822.25 &      &    0.60 \\
                   & logit\_logflip\_b0\_b1 &                     18913.25 &   -0.77 &   -0.80 \\
                   & log\_log\_b1 &                     19493.85 &      &    1.44 \\
                   & logit\_logflip\_b1 &                     19562.58 &      &   -0.54 \\
                   & logit\_logflip\_b0 &                     26426.38 &    1.19 &      \\
                   & logflip\_logflip\_b0 &                     26445.91 &   -1.46 &      \\
                   & logit\_logit\_b0 &                     26519.76 &    1.18 &      \\
                   & log\_log\_b0 &                     26662.65 &   -0.27 &      \\
\bottomrule
\end{tabular}

  \end{tabular}
  }
\end{table*}

Table \ref{table:glm_fit_data} provides parameters fit to calibration functions. For each sample image $x_i$ in the image dataset, define $s_i=f(x_i)$ to be the score (the output of the top-scoring logit after softmax) and $y_i \in \{0,1\}$ to be the classification ($y_i=1$ when the top-scoring logit correctly classified image $x_i$) for the sample image. The loss for the binary generalized linear model (GLM) across different combinations of link functions $g(y)$ and transform functions $t(s)$ was optimized via the standard loss 
(Gelman et al., 2004):
\begin{equation}
    \argmin_{b_0, b_1} \sum_i -\ln p_i^{y_i}(1-p_i)^{1-y_i}, \quad p_i=g^{-1}(b_0 + b_1 t(s_i))
\end{equation}
For each dataset, the GLM of best fit was selected via the Akaike Information Criteria using the likelihood at the optimized parameter values. 

\subsection{Comparing \texorpdfstring{\ecebin{}}{ECEbin} computed on simulated data versus real data}

In Figure \ref{fig:beta_glm_fit}c, we compare the \ecebin{} computed on the original logit output of each model to the average \ecebin{} we obtain after sampling $1{,}000$ simulated datasets from our parametric fits. Table \ref{table:eece_sece} reports the \ecebin{} measurements that we plot in Figure \ref{fig:beta_glm_fit}.
We observe that the two measurements of \ecebin{} are relatively close, indicating that our parametric models are well-fit to the original data.

\begin{table}[ht!]
\centering
\caption{\ecebin{} reported in Figure \ref{fig:beta_glm_fit}c.}
\label{table:eece_sece}
\begin{tabular}{lrr}
\toprule
{} &  \ecebin{} (\%) &  <\ecebin{}> (\%, simulated) \\
\midrule
resnet110\_c10      &     6.67 &     8.42 \\
resnet110\_SD\_c10   &     6.54 &     8.79 \\
resnet\_wide32\_c10  &     6.09 &     8.44 \\
densenet40\_c10     &     6.70 &     8.09 \\
resnet110\_c100     &    20.26 &    18.87 \\
resnet110\_SD\_c100  &    17.44 &    15.78 \\
resnet\_wide32\_c100 &    20.40 &    17.53 \\
densenet40\_c100    &    23.12 &    19.69 \\
resnet152\_imgnet   &     6.85 &     9.26 \\
densenet161\_imgnet &     6.15 &     6.87 \\
\bottomrule
\end{tabular}
\end{table}

\newpage
\clearpage
\section{Bias and variance in calibration metrics}
\label{apx:bias_and_variance}

\subsection{Bias}
We evaluate bias for various calibration metrics using both equal-width and equal-mass binning as we vary both the sample size $n$ and the number of bins $b$.  These plots should be seen as an alternative visualization to \ref{fig:ml_bias} where we additionally compare to different choices for the fixed number of bins $b$.  
Since the \cesweep{} metrics adaptively choose a different number of bins for each sample size, we display the bin number for this metric as $-1$.

We find that \ecebin{} can overestimate the true calibration error and there exists an optimal number of bins that produces the least biased estimator that changes with the number of samples $n$. 
Additionally, equal mass binning generally results in a less biased metric than equal width binning.

\begin{figure}[ht!]
    \centering
    \begin{subfigure}{0.25\textwidth}
      \includegraphics[width=0.9\linewidth]{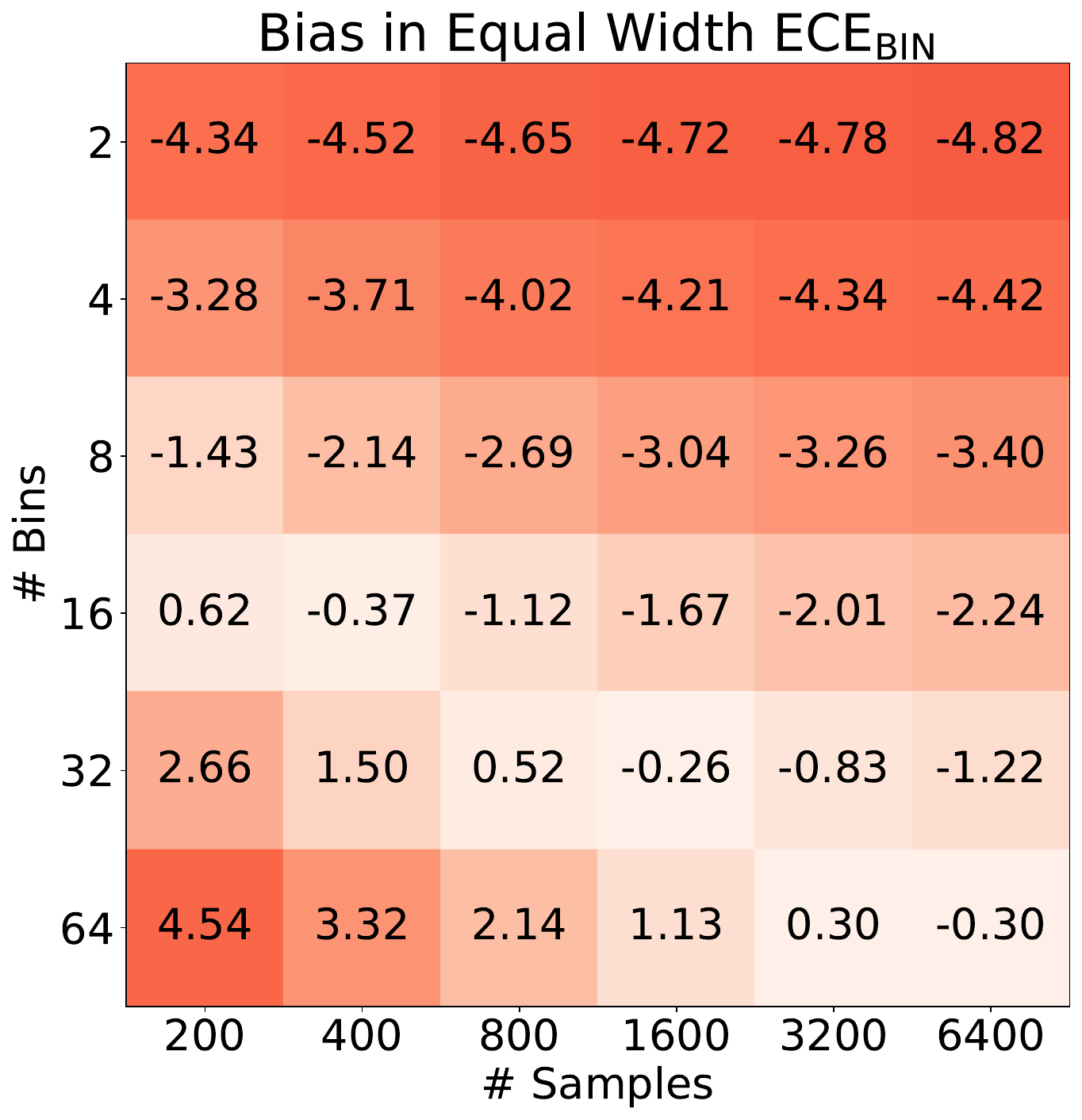}
    \end{subfigure}
    \begin{subfigure}{0.25\textwidth}
      \includegraphics[width=0.9\linewidth]{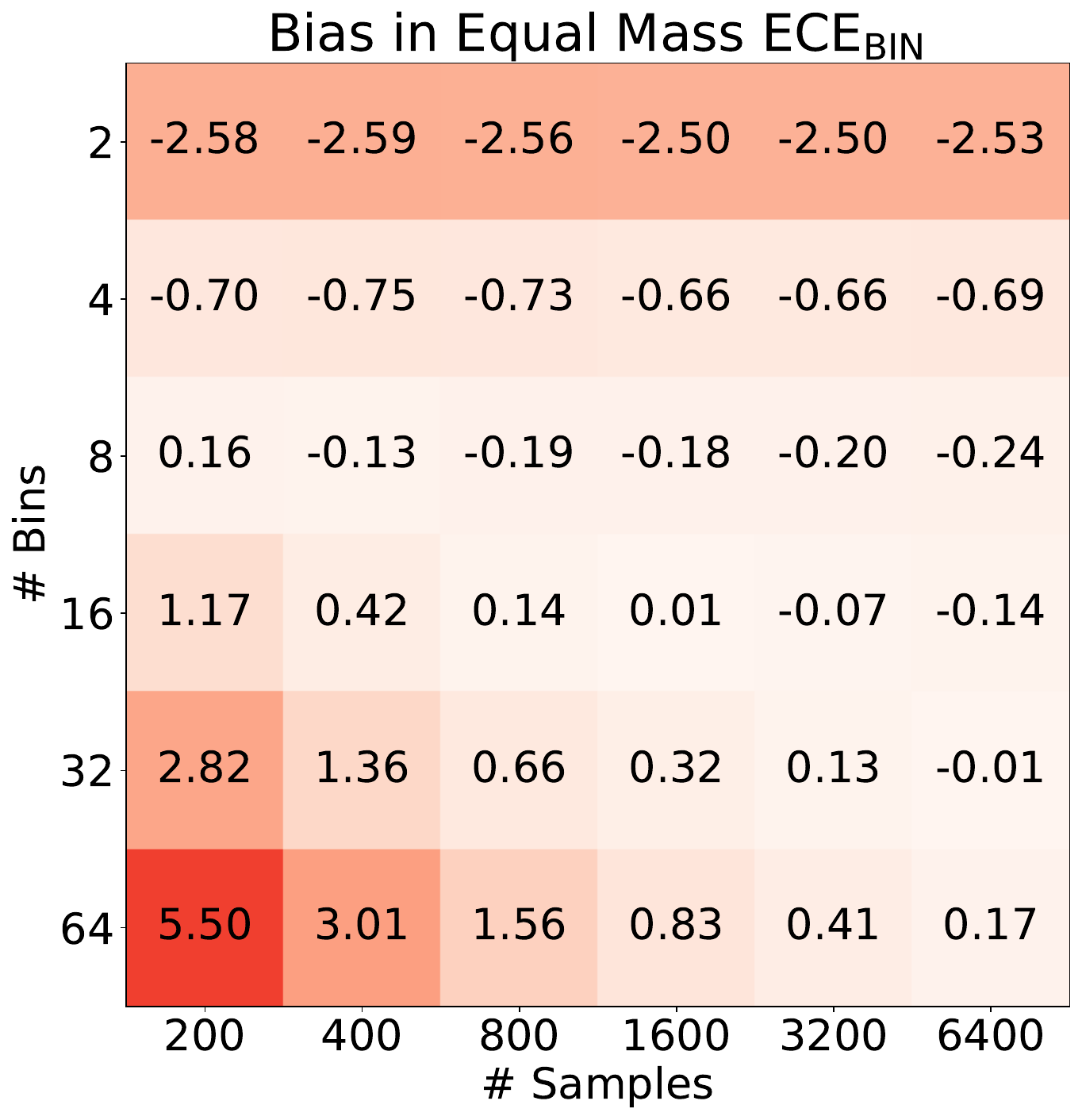}
    \end{subfigure}
    \begin{subfigure}{0.25\textwidth}
      \includegraphics[width=0.9\linewidth]{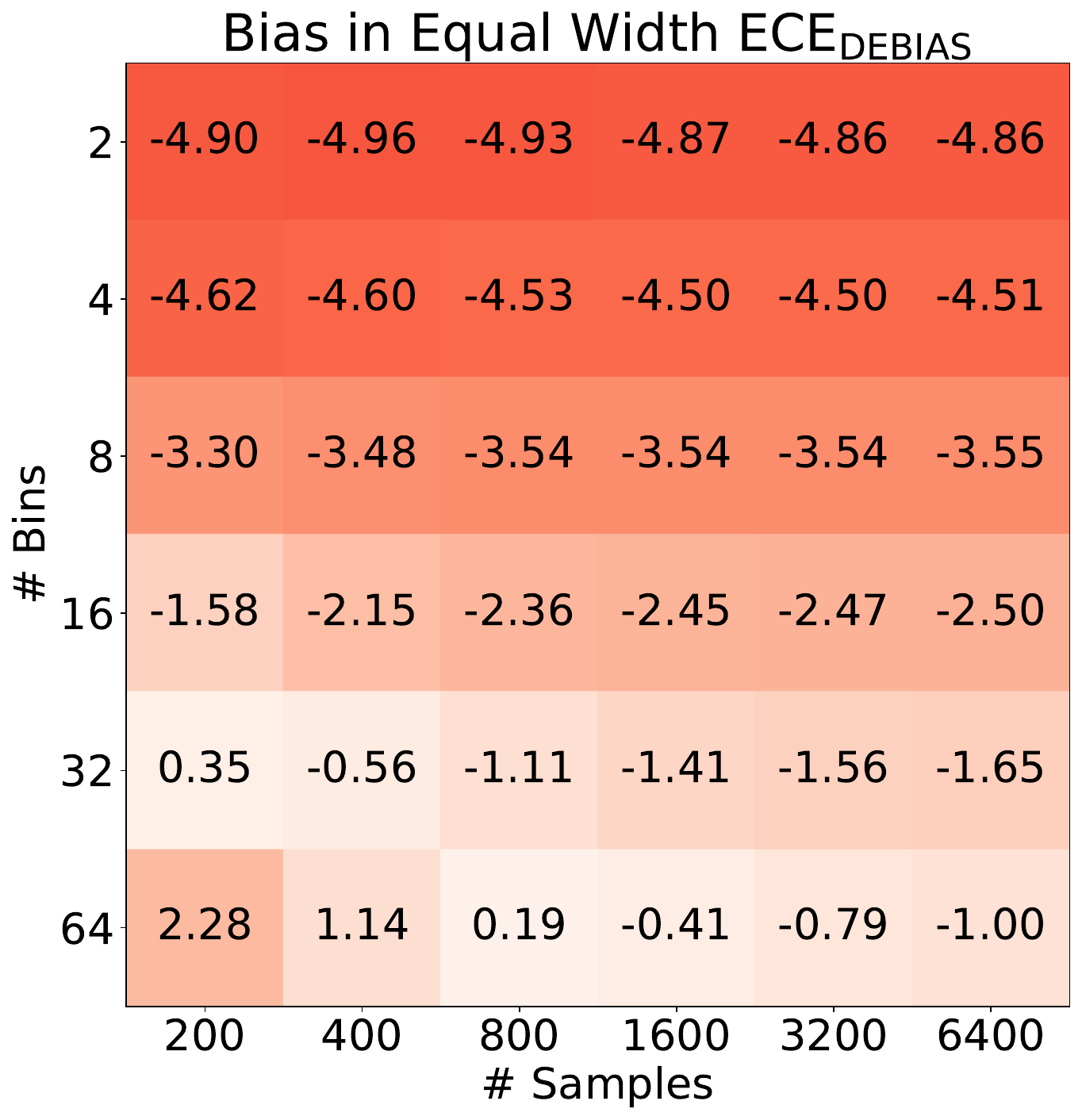}
    \end{subfigure}
    \begin{subfigure}{0.25\textwidth}
      \includegraphics[width=0.9\linewidth]{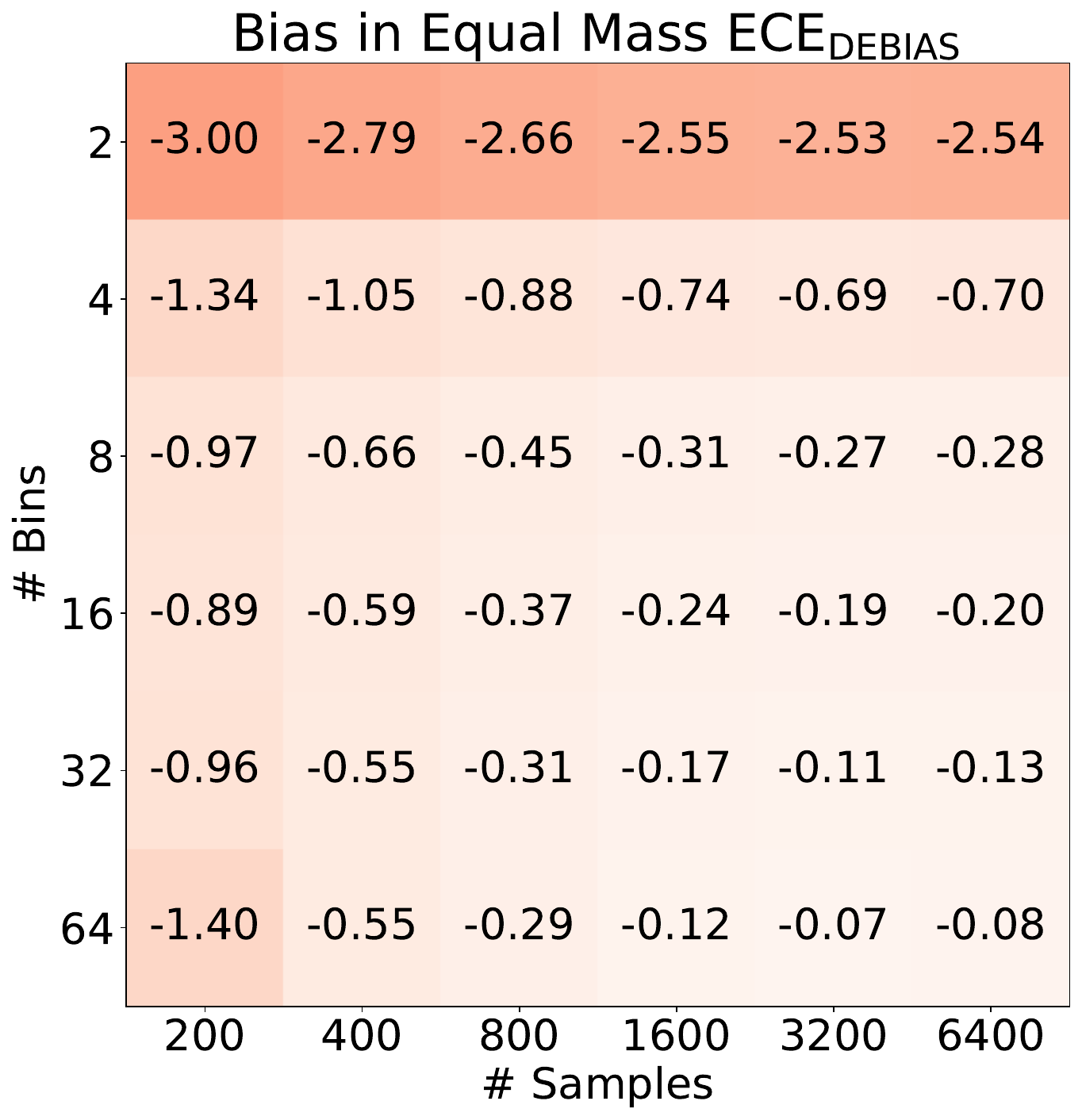}
    \end{subfigure}
    \begin{subfigure}{0.25\textwidth}
      \includegraphics[width=0.9\linewidth]{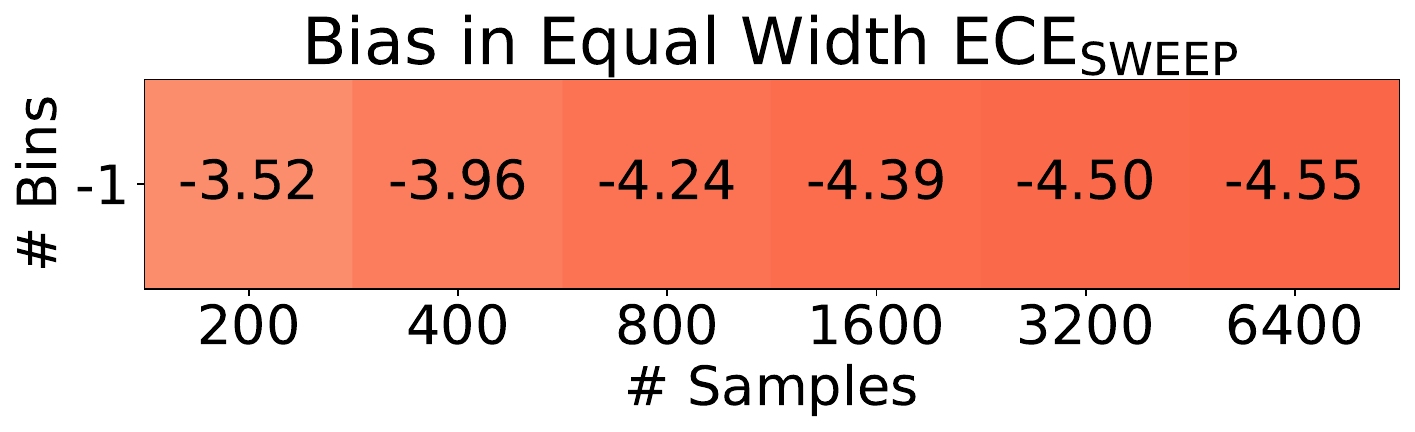}
    \end{subfigure}
    \begin{subfigure}{0.25\textwidth}
      \includegraphics[width=0.9\linewidth]{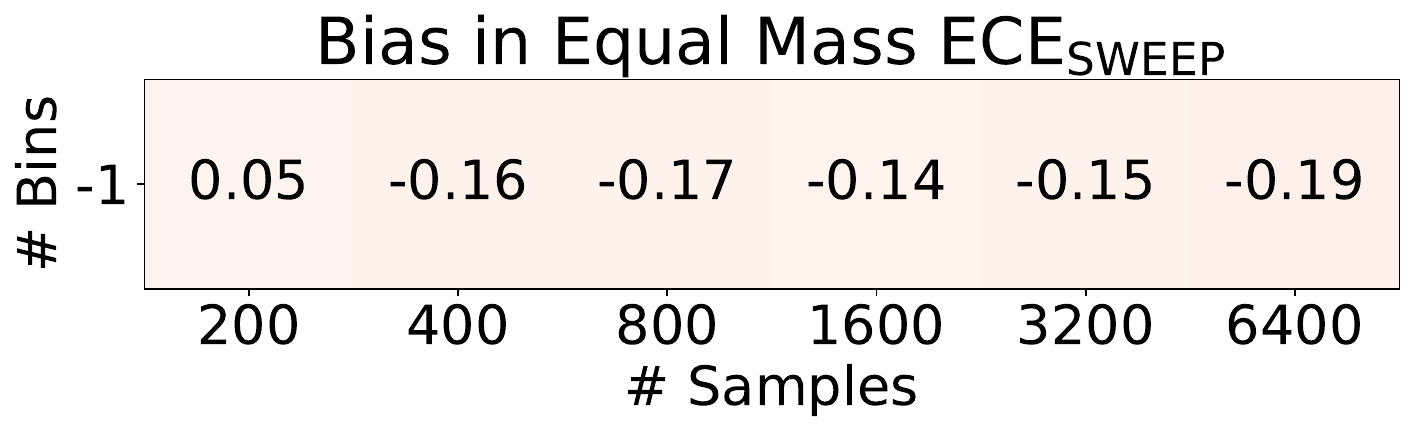}
    \end{subfigure}
    
  \caption{\textbf{Bias for various calibration metrics assuming curves fit to CIFAR-10 ResNet-110 output.}
  We plot bias for various calibration metrics using both equal-width binning (left column) and equal-mass binning (right column) as we vary both the sample size $n$ and the number of bins $b$. 
  }
  \label{fig:bias_resnet110_c10}
\end{figure}

\textbf{CIFAR-10 ResNet-110.}
Figure \ref{fig:bias_resnet110_c10} assume parametric curves for $p(f(x))$ and $\truecurve$ that we obtain from maximum-likelihood fits to CIFAR-10 ResNet-110 model output.  

\begin{figure}[ht!]
    \centering
    \begin{subfigure}{0.25\textwidth}
      \includegraphics[width=0.9\linewidth]{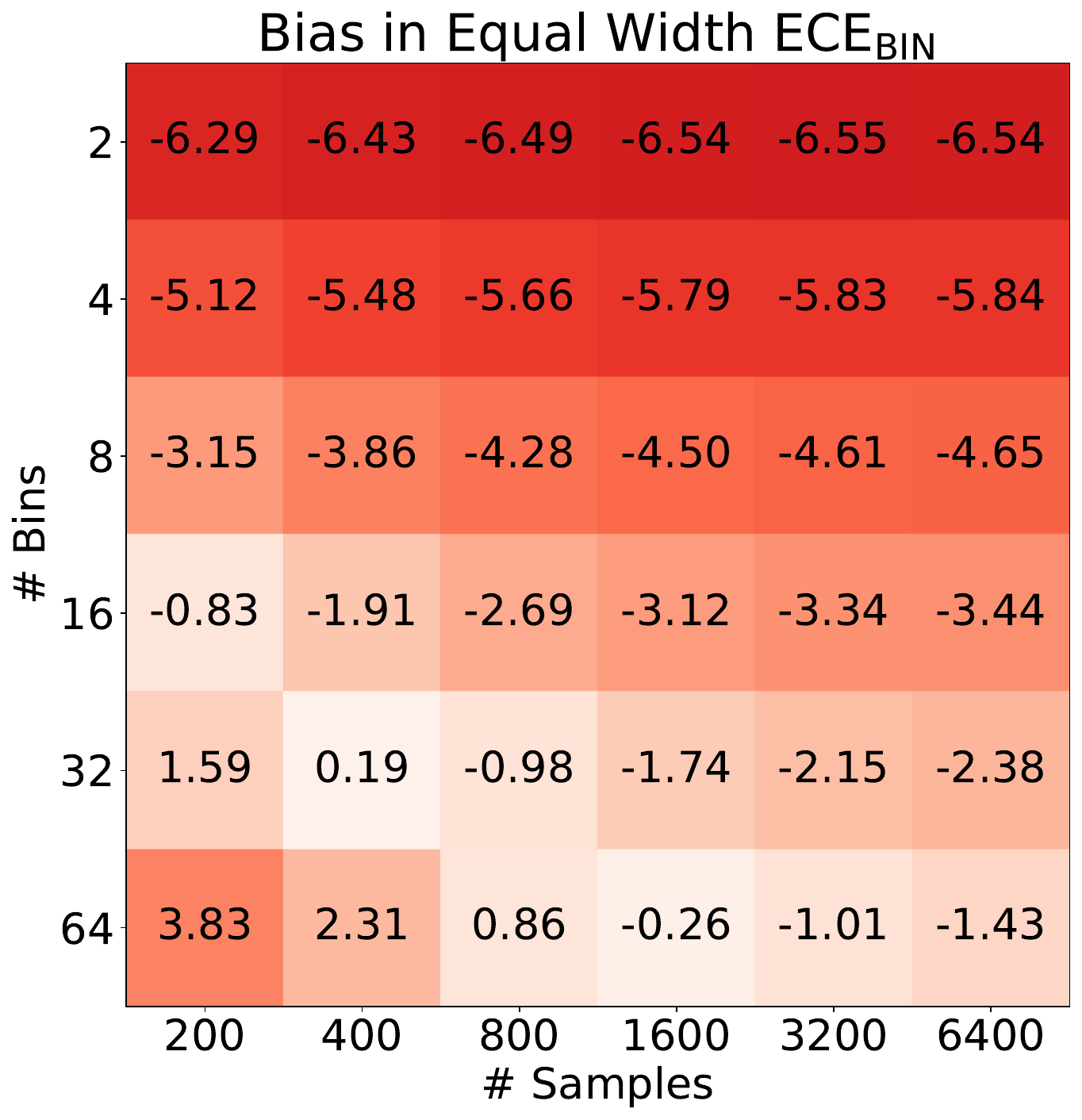}
    \end{subfigure}
    \begin{subfigure}{0.25\textwidth}
      \includegraphics[width=0.9\linewidth]{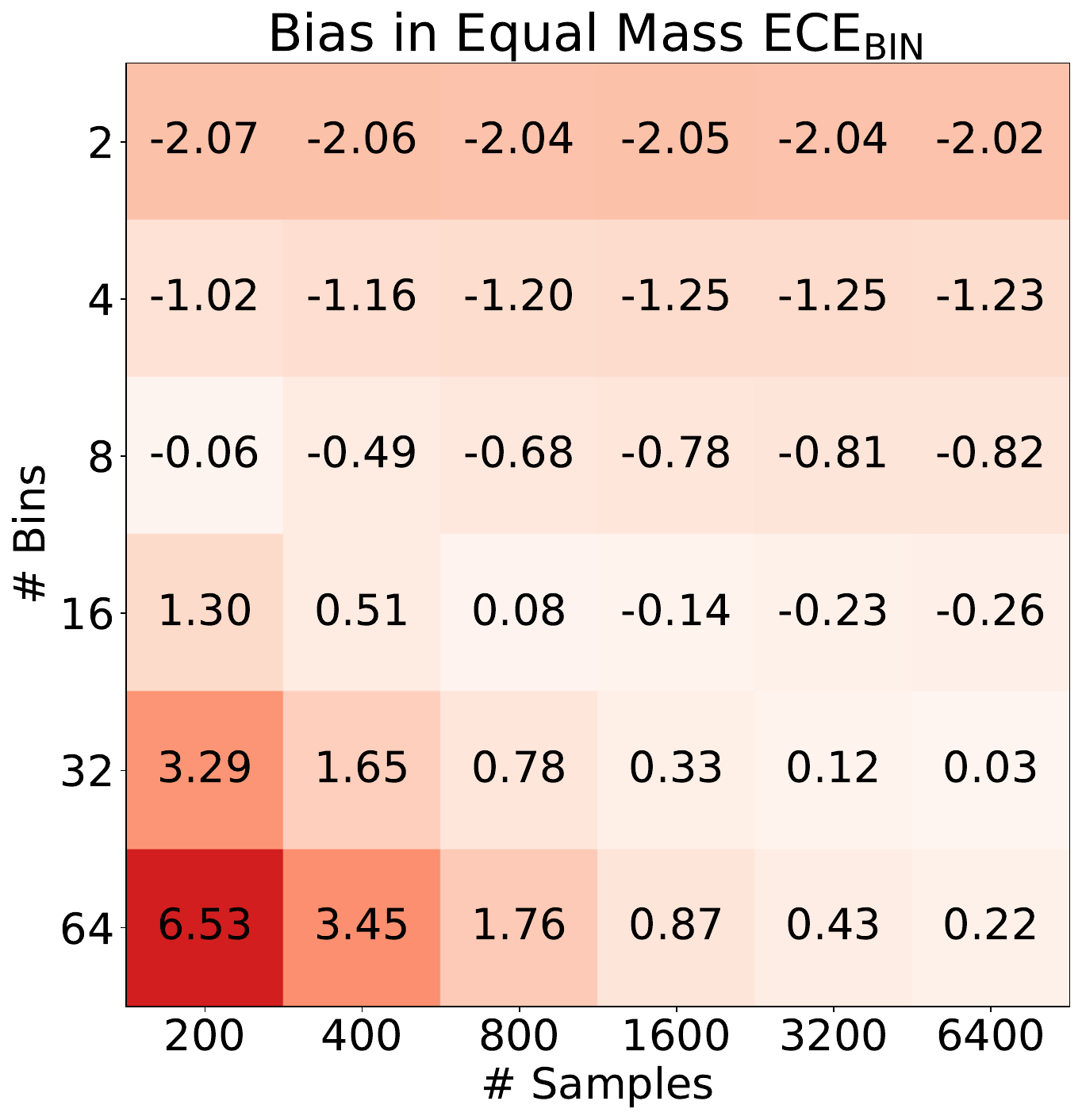}
    \end{subfigure}
    \begin{subfigure}{0.25\textwidth}
      \includegraphics[width=0.9\linewidth]{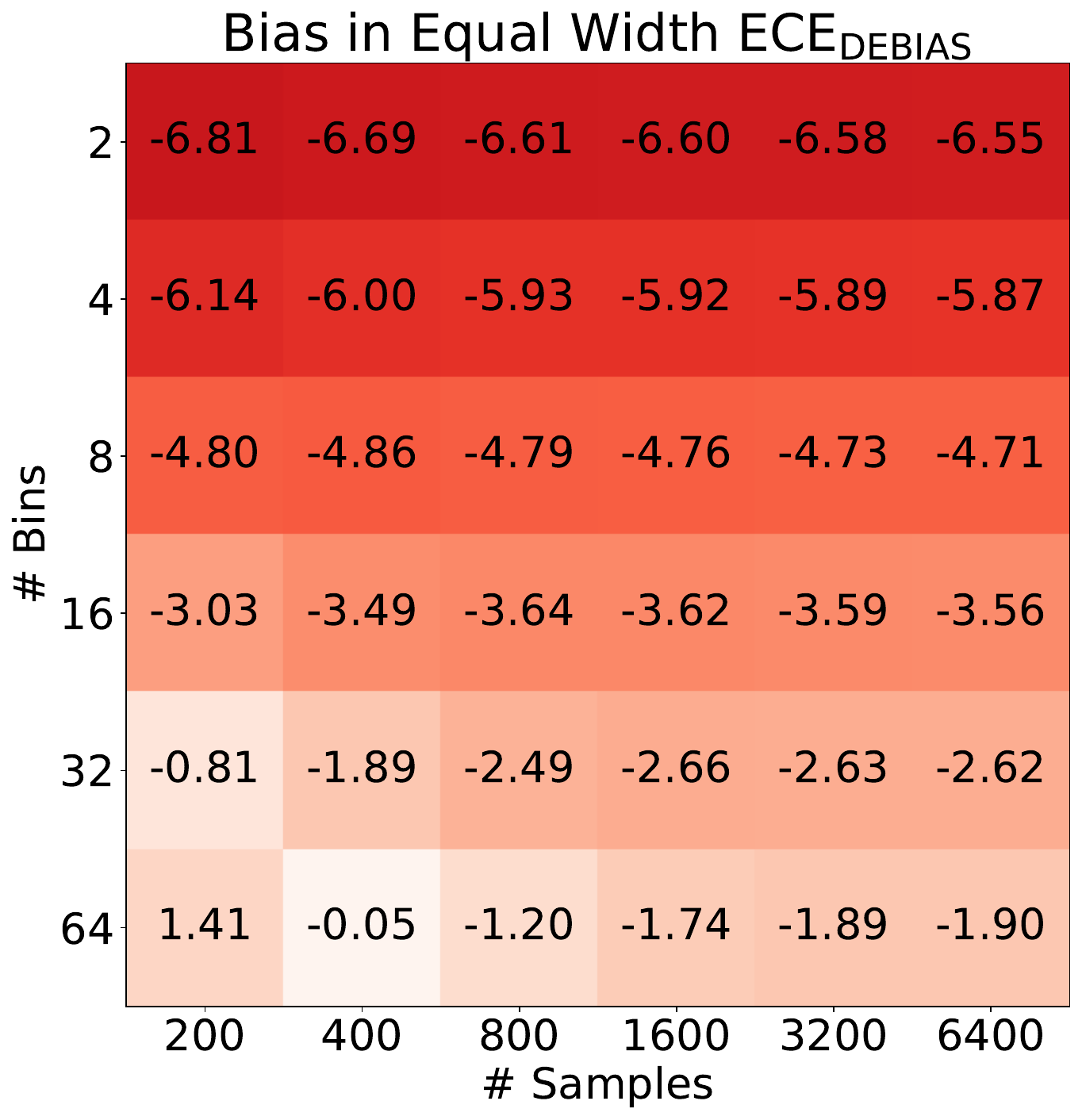}
    \end{subfigure}
    \begin{subfigure}{0.25\textwidth}
      \includegraphics[width=0.9\linewidth]{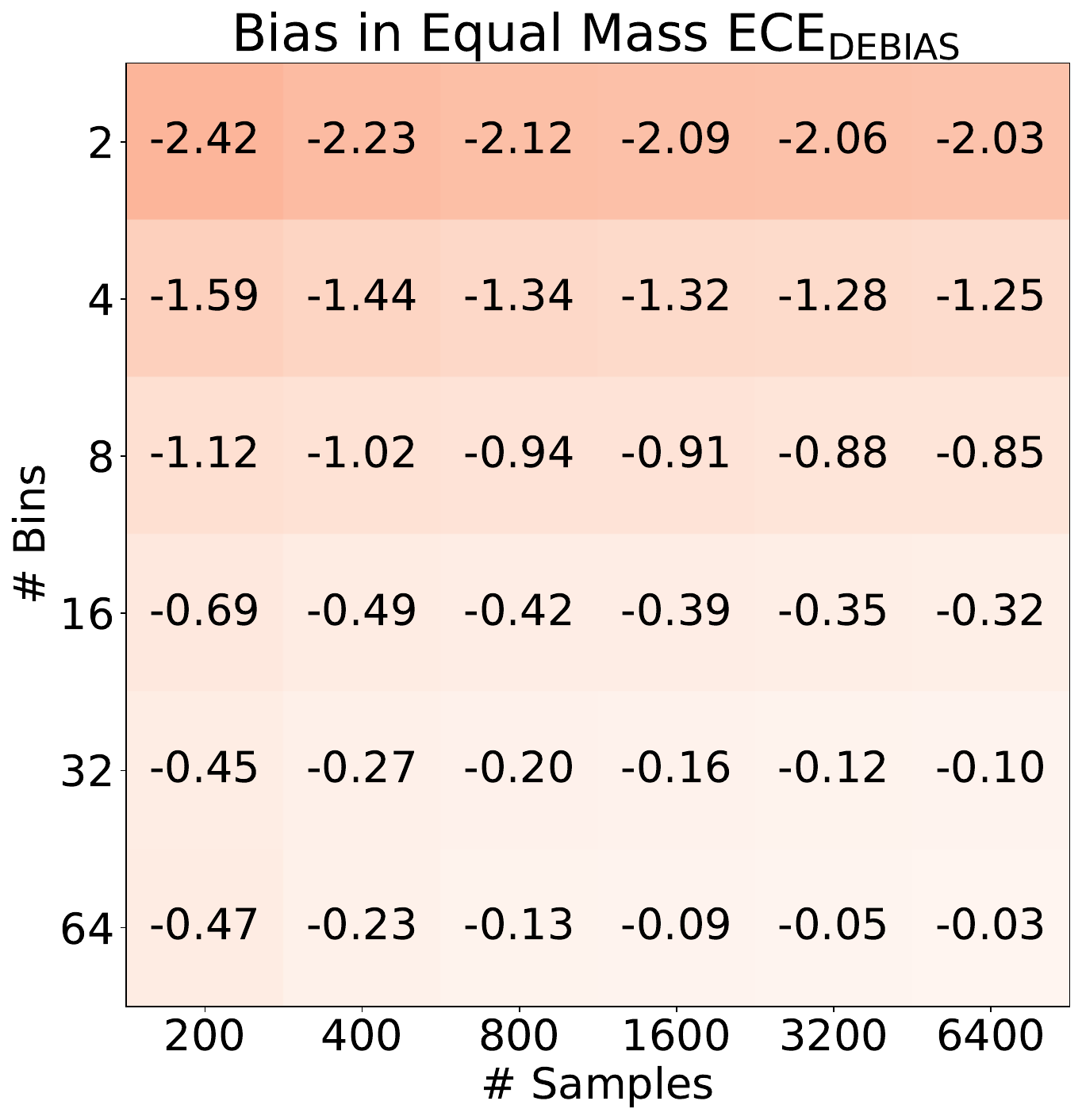}
    \end{subfigure}
    \begin{subfigure}{0.25\textwidth}
      \includegraphics[width=0.9\linewidth]{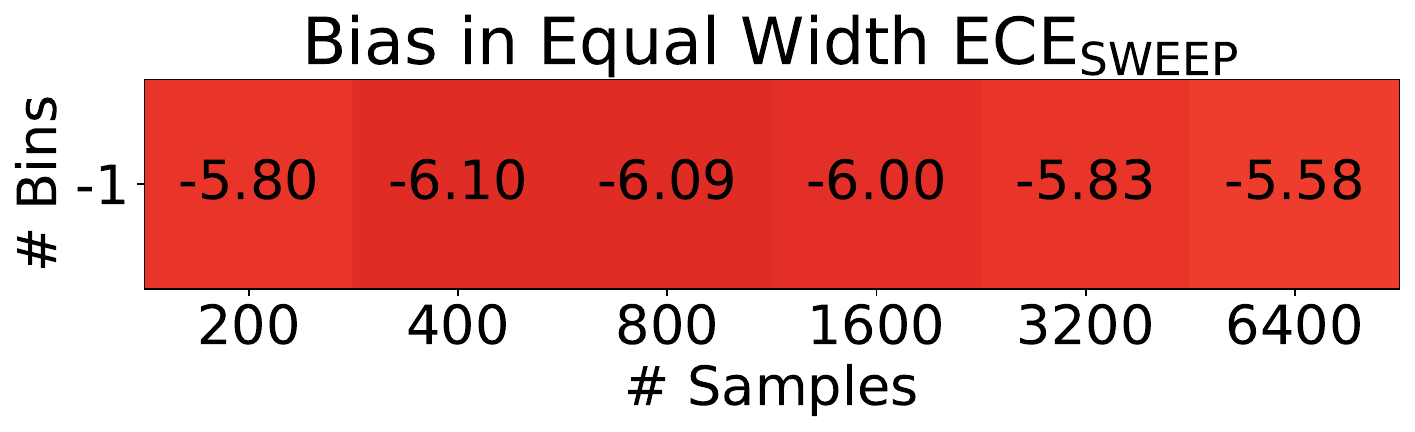}
    \end{subfigure}
    \begin{subfigure}{0.25\textwidth}
      \includegraphics[width=0.9\linewidth]{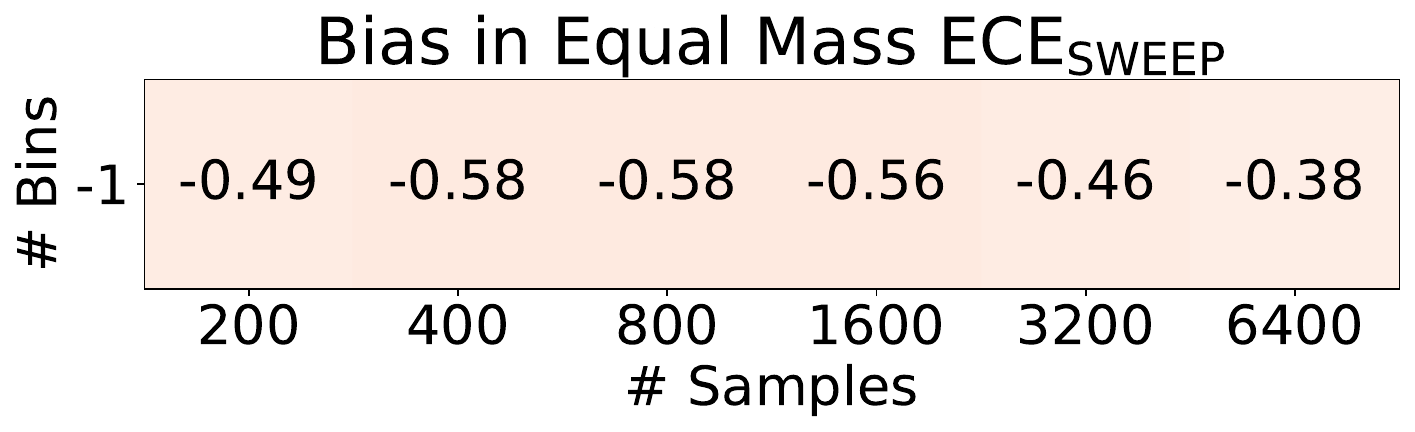}
    \end{subfigure}
    
  \caption{\textbf{Bias for various calibration metrics assuming curves fit to CIFAR-100 Wide ResNet-32 output.}
  We plot bias for various calibration metrics using both equal-width binning (left column) and equal-mass binning (right column) as we vary both the sample size $n$ and the number of bins $b$. 
  }
  \label{fig:bias_resnet_wide32_c100}
\end{figure}

\textbf{CIFAR-100 Wide ResNet-32.}
Figure \ref{fig:bias_resnet_wide32_c100} assume parametric curves for $p(f(x))$ and $\truecurve$ that we obtain from maximum-likelihood fits to CIFAR-100 Wide ResNet-32 model output.  

\begin{figure}[ht!]
    \centering
    \begin{subfigure}{0.25\textwidth}
      \includegraphics[width=0.9\linewidth]{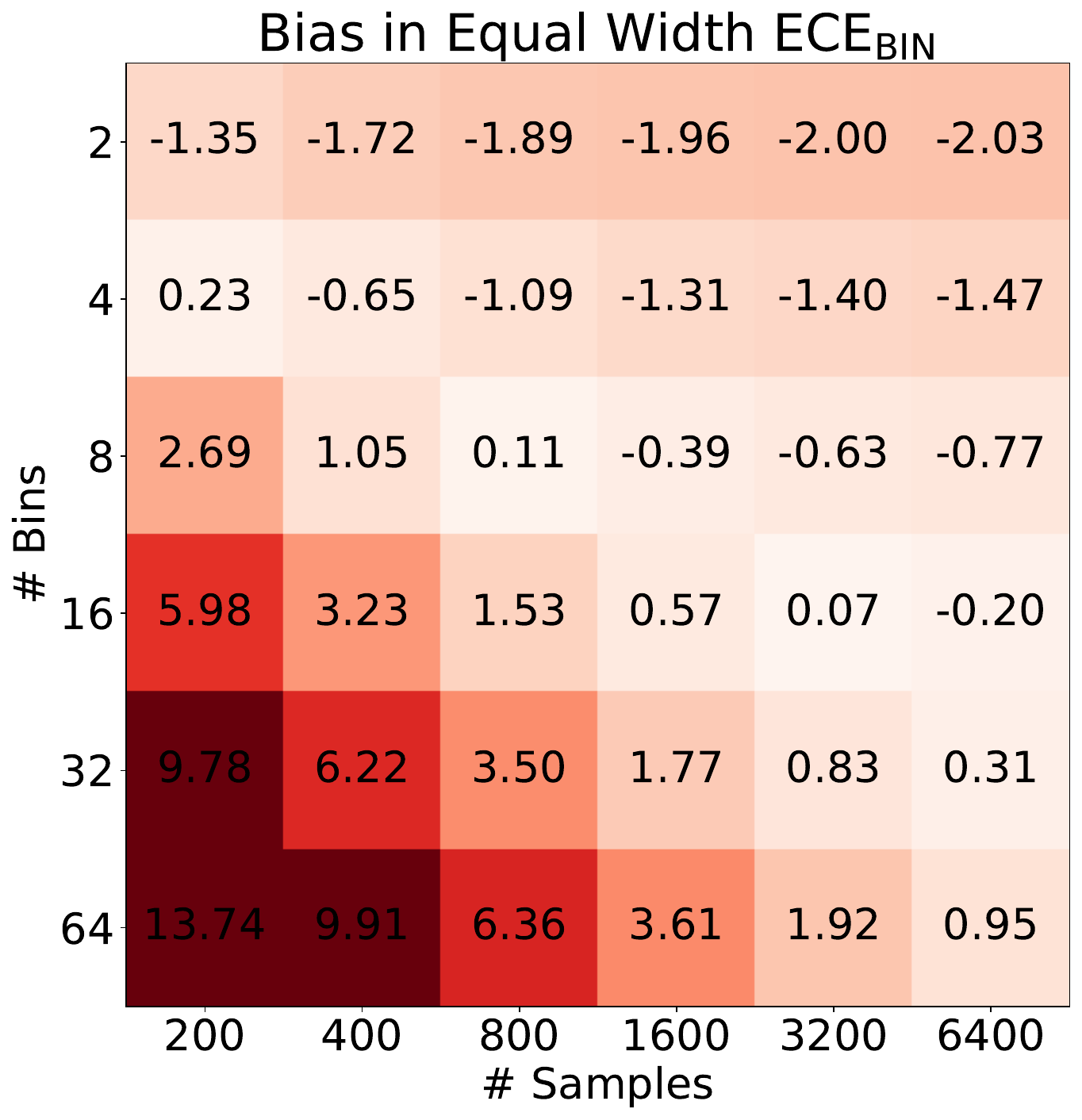}
    \end{subfigure}
    \begin{subfigure}{0.25\textwidth}
      \includegraphics[width=0.9\linewidth]{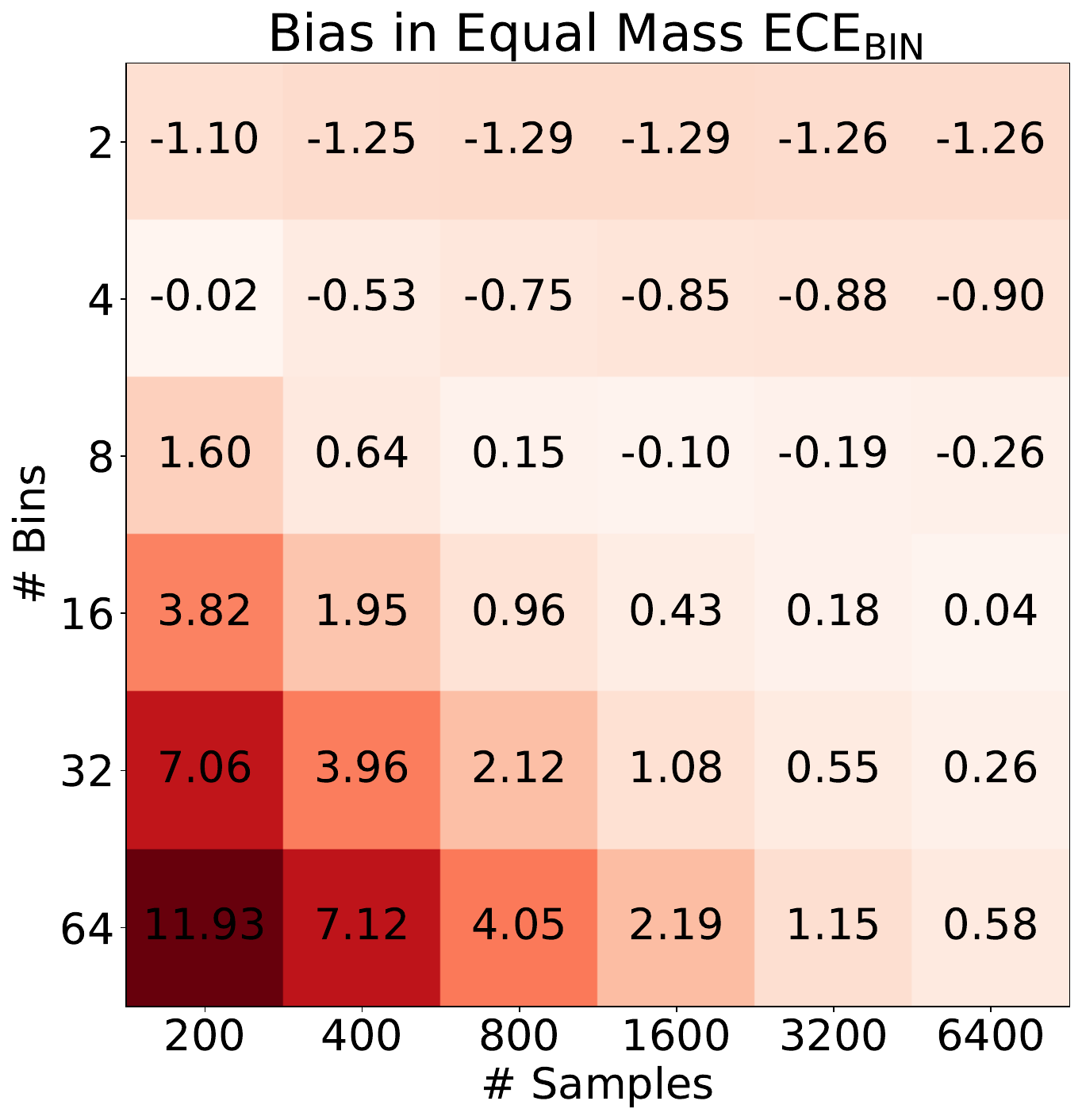}
    \end{subfigure}
    \begin{subfigure}{0.25\textwidth}
      \includegraphics[width=0.9\linewidth]{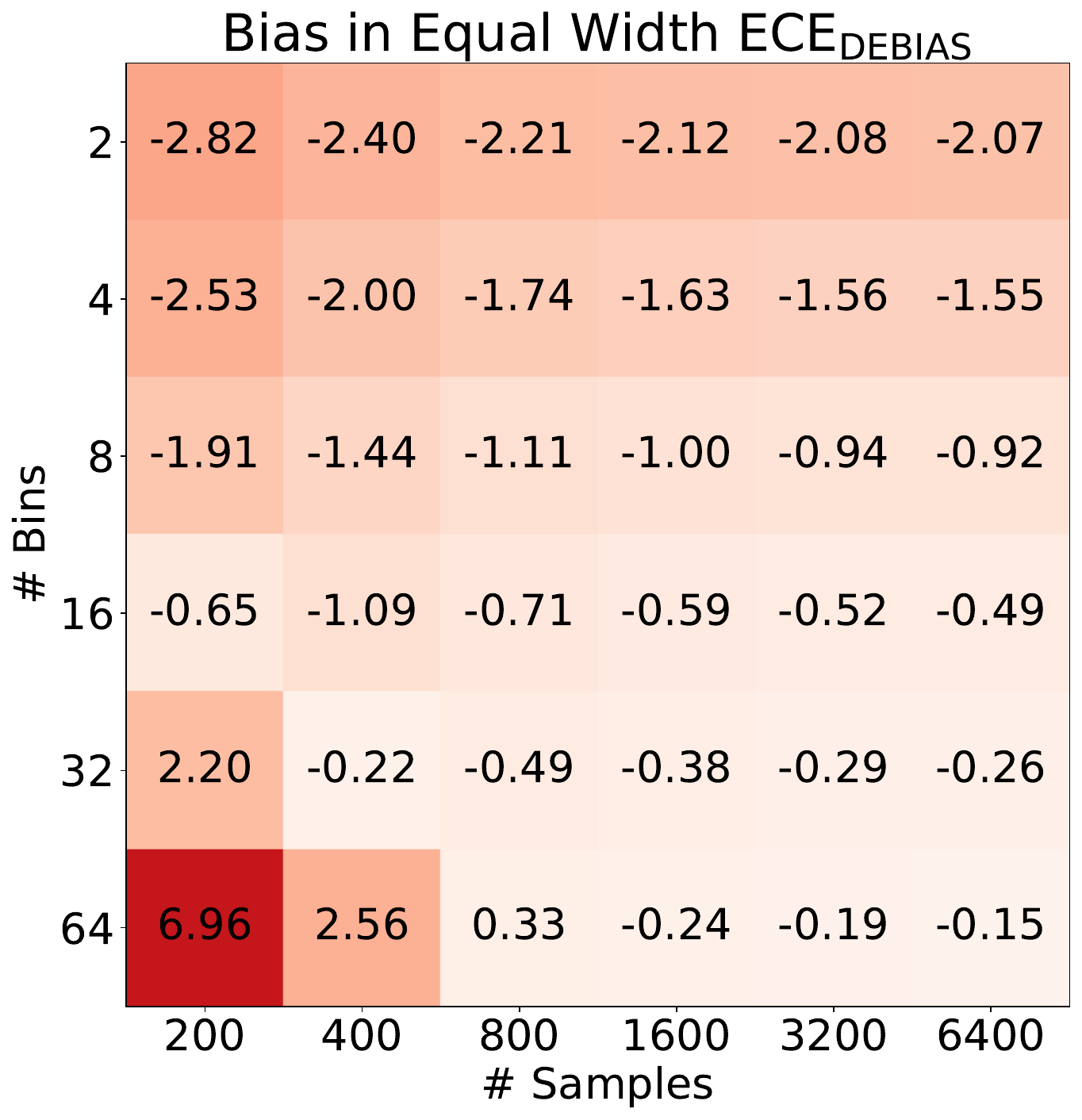}
    \end{subfigure}
    \begin{subfigure}{0.25\textwidth}
      \includegraphics[width=0.9\linewidth]{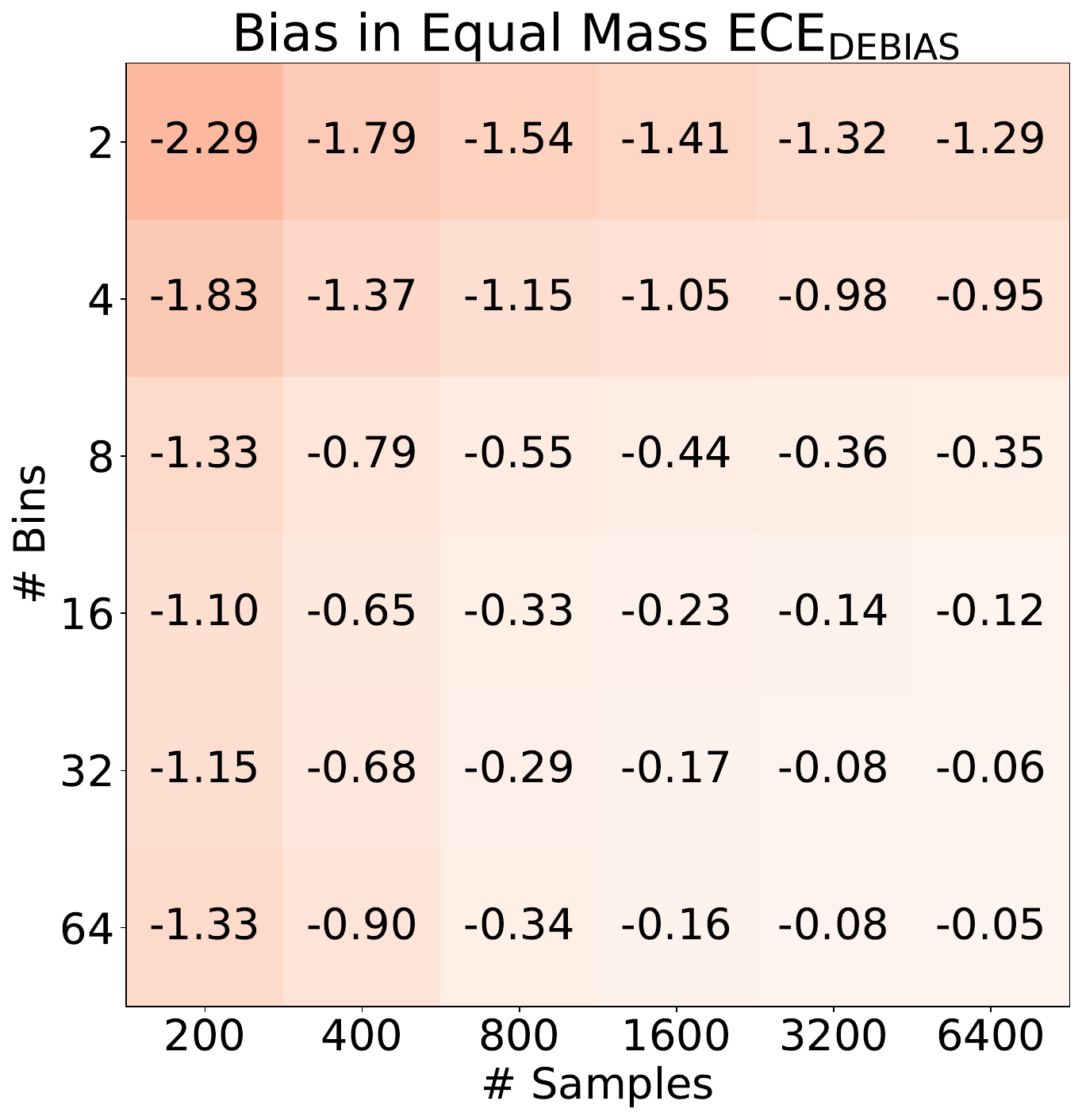}
    \end{subfigure}
    \begin{subfigure}{0.25\textwidth}
      \includegraphics[width=0.9\linewidth]{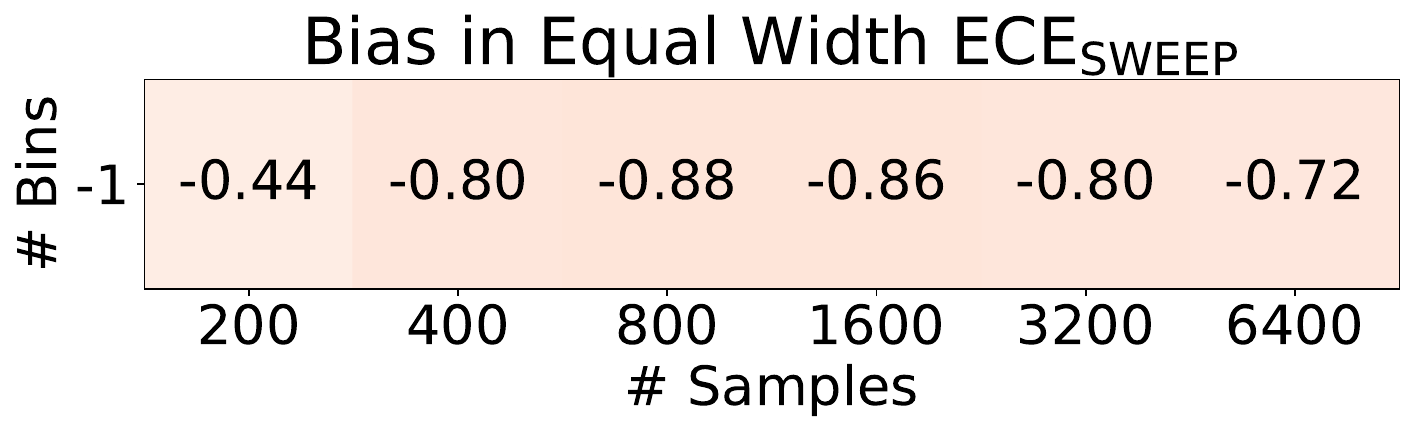}
    \end{subfigure}
    \begin{subfigure}{0.25\textwidth}
      \includegraphics[width=0.9\linewidth]{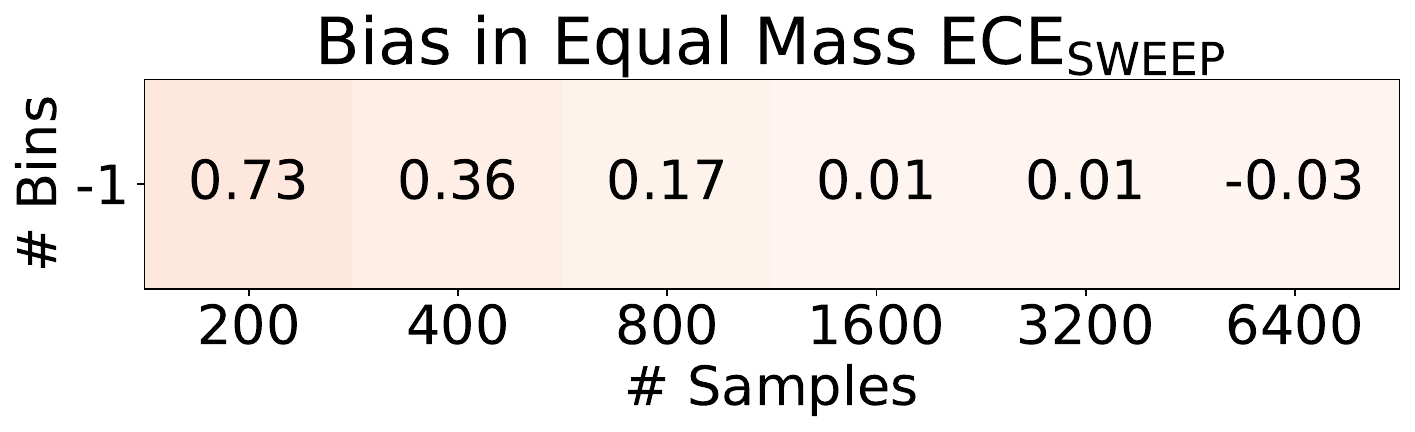}
    \end{subfigure}
    
  \caption{\textbf{Bias for various calibration metrics assuming curves fit to ImageNet ResNet-152 output.}
  We plot bias for various calibration metrics using both equal-width binning (left column) and equal-mass binning (right column) as we vary both the sample size $n$ and the number of bins $b$. 
  }
  \label{fig:bias_resnet152_imgnet}
\end{figure}

\textbf{ImageNet ResNet-152.}
Figure \ref{fig:bias_resnet152_imgnet} assume parametric curves for $p(f(x))$ and $\truecurve$ that we obtain from maximum-likelihood fits to ImageNet ResNet-152 model output.  

\clearpage
\newpage
\subsection{Variance}
\label{apx:variance}
We also compute the \textit{variance} for various calibration metrics using both equal-width and equal-mass binning as we vary both the sample size $n$ and the number of bins $b$. 
As expected, the variance decreases with number of samples, but, unlike the bias, there is no clear dependence on the number of bins.

\begin{figure}[ht!]
    \centering
    \begin{subfigure}{0.25\textwidth}
      \includegraphics[width=0.9\linewidth]{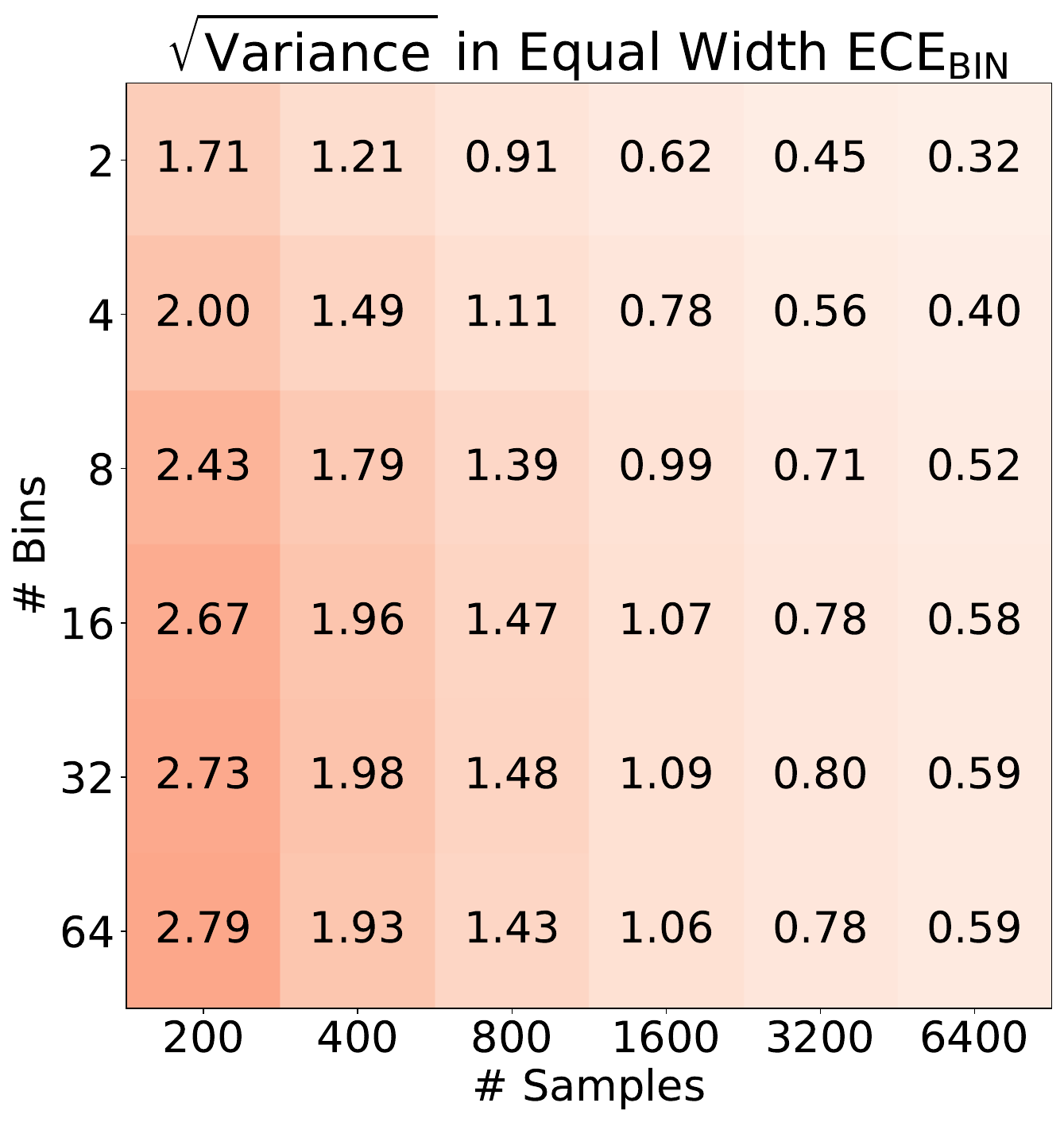}
    \end{subfigure}
    \begin{subfigure}{0.25\textwidth}
      \includegraphics[width=0.9\linewidth]{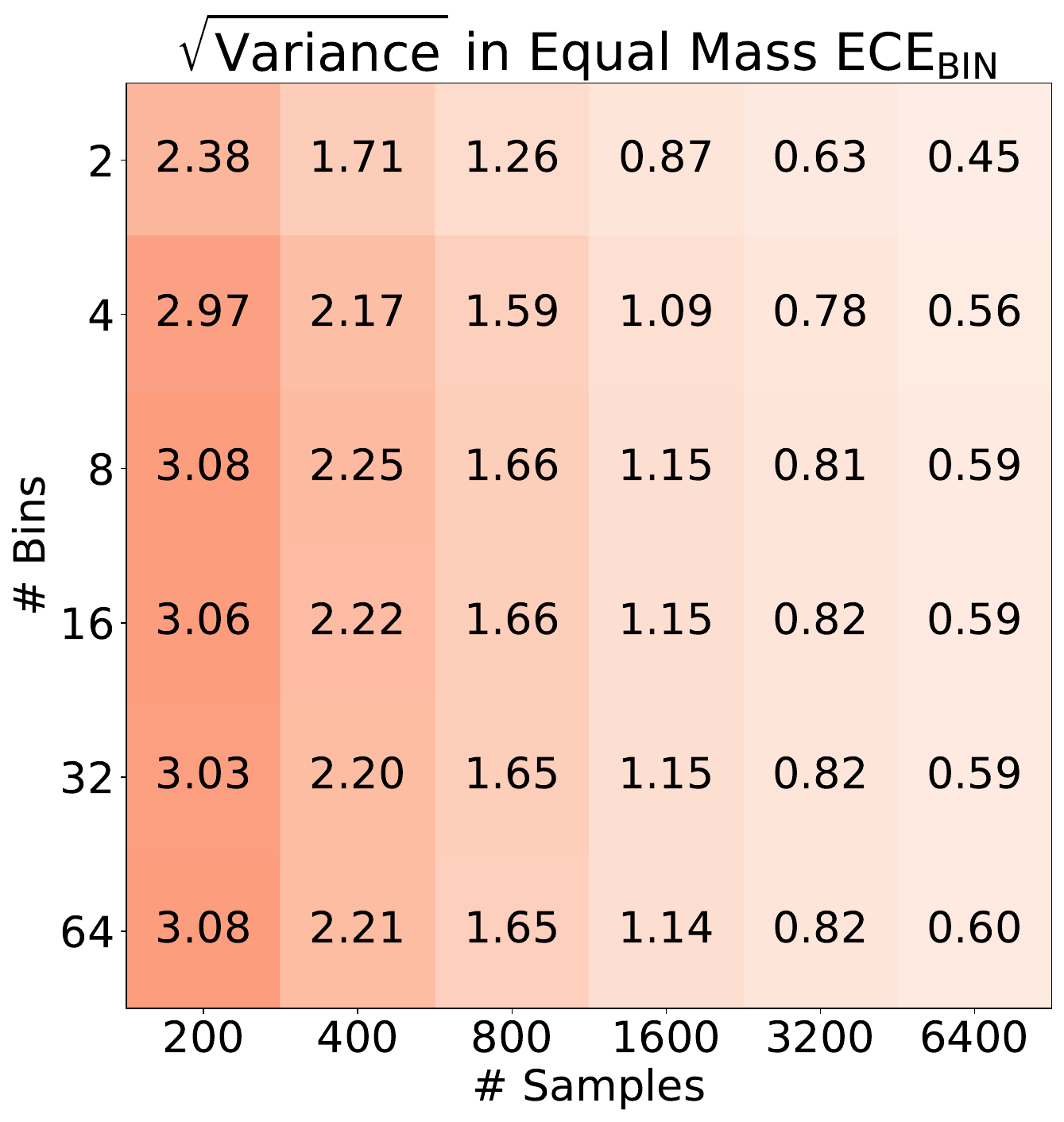}
    \end{subfigure}
    \begin{subfigure}{0.25\textwidth}
      \includegraphics[width=0.9\linewidth]{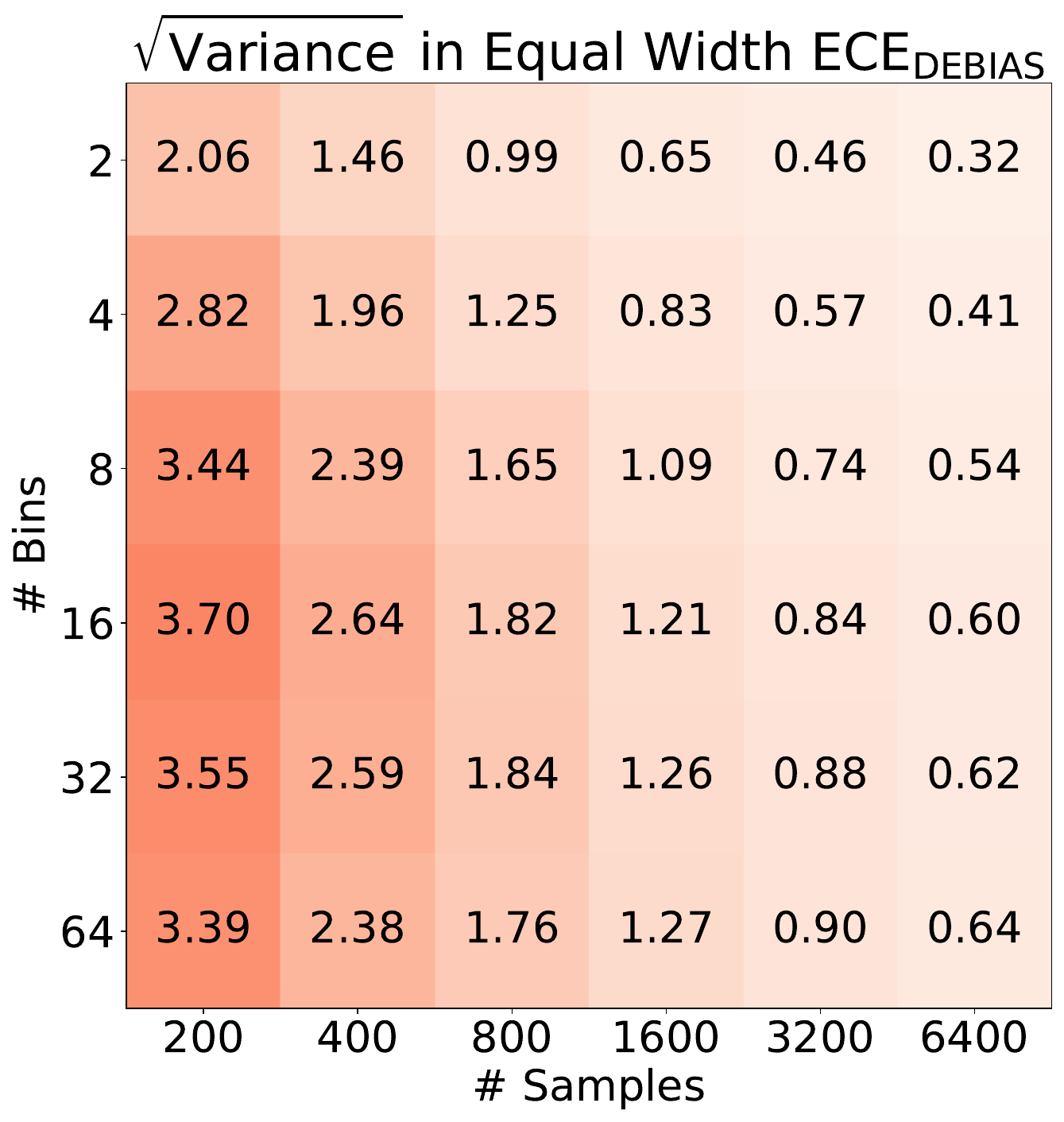}
    \end{subfigure}
    \begin{subfigure}{0.25\textwidth}
      \includegraphics[width=0.9\linewidth]{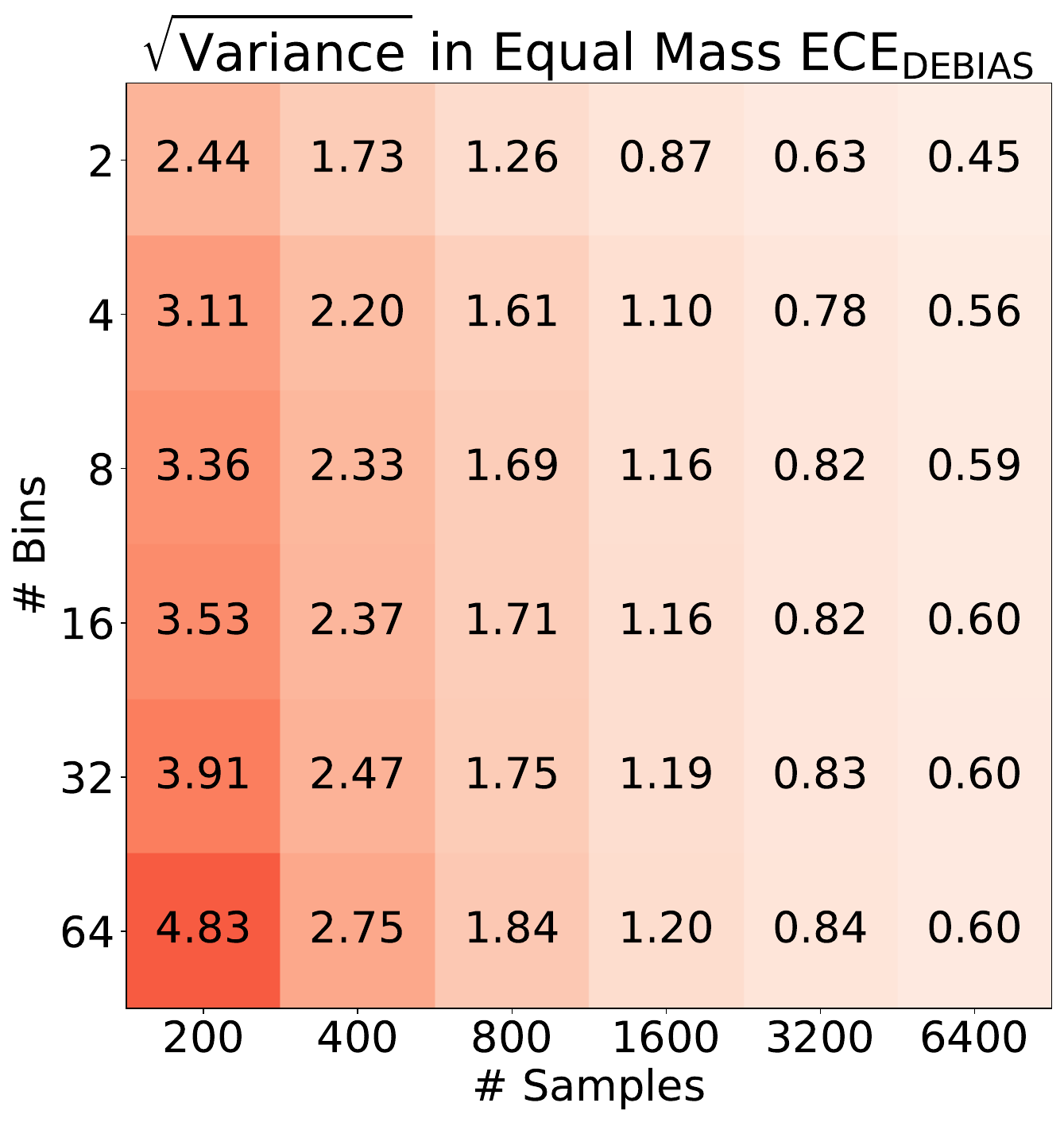}
    \end{subfigure}
    \begin{subfigure}{0.25\textwidth}
      \includegraphics[width=0.9\linewidth]{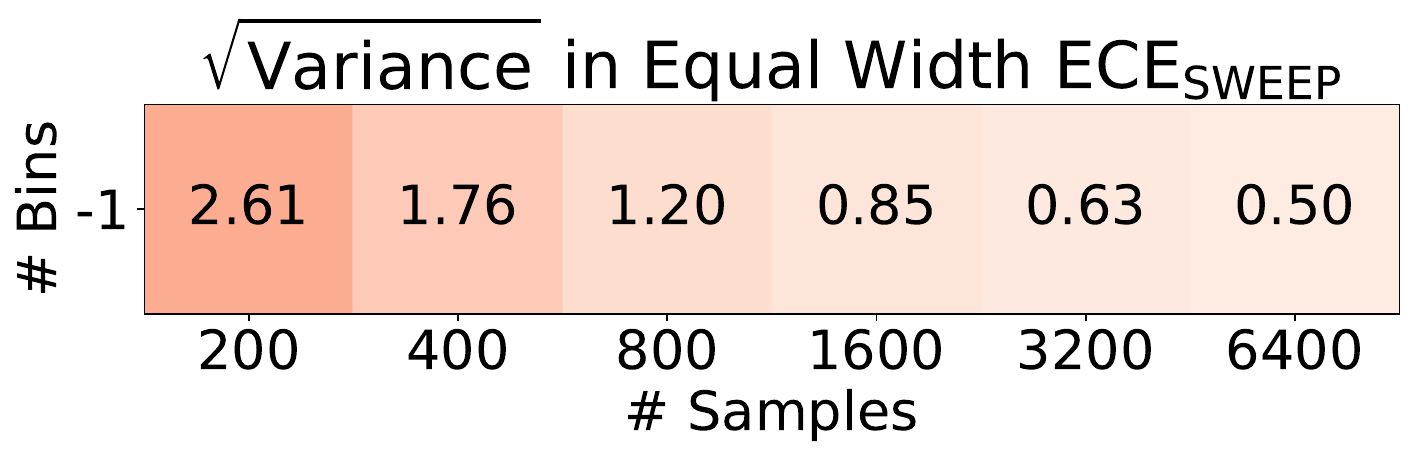}
    \end{subfigure}
    \begin{subfigure}{0.25\textwidth}
      \includegraphics[width=0.9\linewidth]{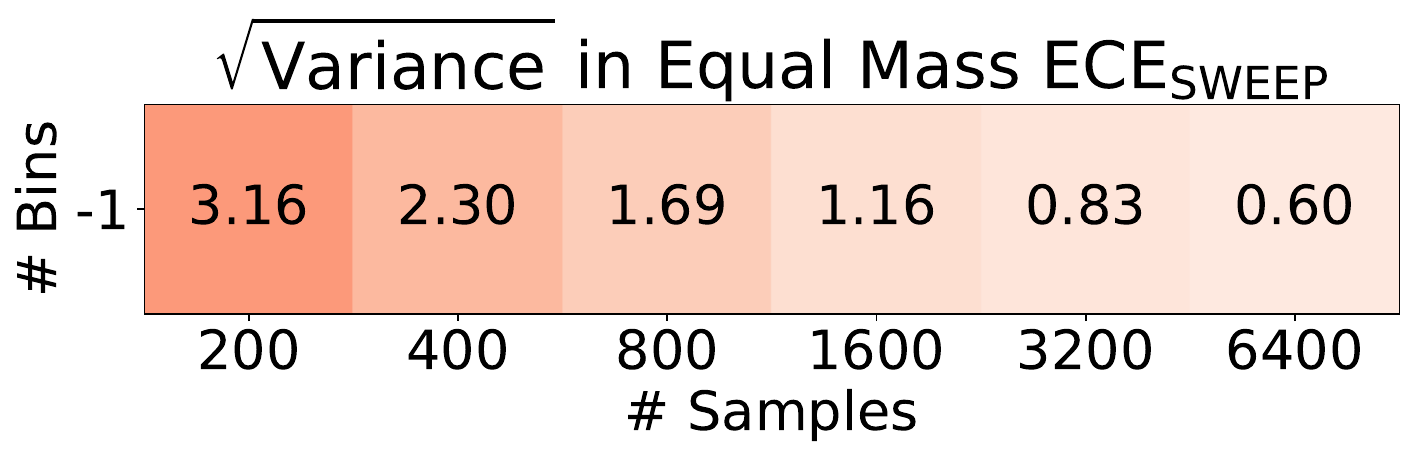}
    \end{subfigure}
    
  \caption{\textbf{$\sqrt{\text{Variance}}$ for various calibration metrics assuming curves fit to CIFAR-10 ResNet-110 output.}
  We plot $\sqrt{\text{Variance}}$ for various calibration metrics using both equal-width binning (left column) and equal-mass binning (right column) as we vary both the sample size $n$ and the number of bins $b$. 
  }
  \label{fig:variance_resnet110_c10}
\end{figure}

\textbf{CIFAR-10 ResNet-110.}
Figure \ref{fig:variance_resnet110_c10} assume parametric curves for $p(f(x))$ and $\truecurve$ that we obtain from maximum-likelihood fits to CIFAR-10 ResNet-110 model output.  

\begin{figure}[ht!]
    \centering
    \begin{subfigure}{0.25\textwidth}
      \includegraphics[width=0.9\linewidth]{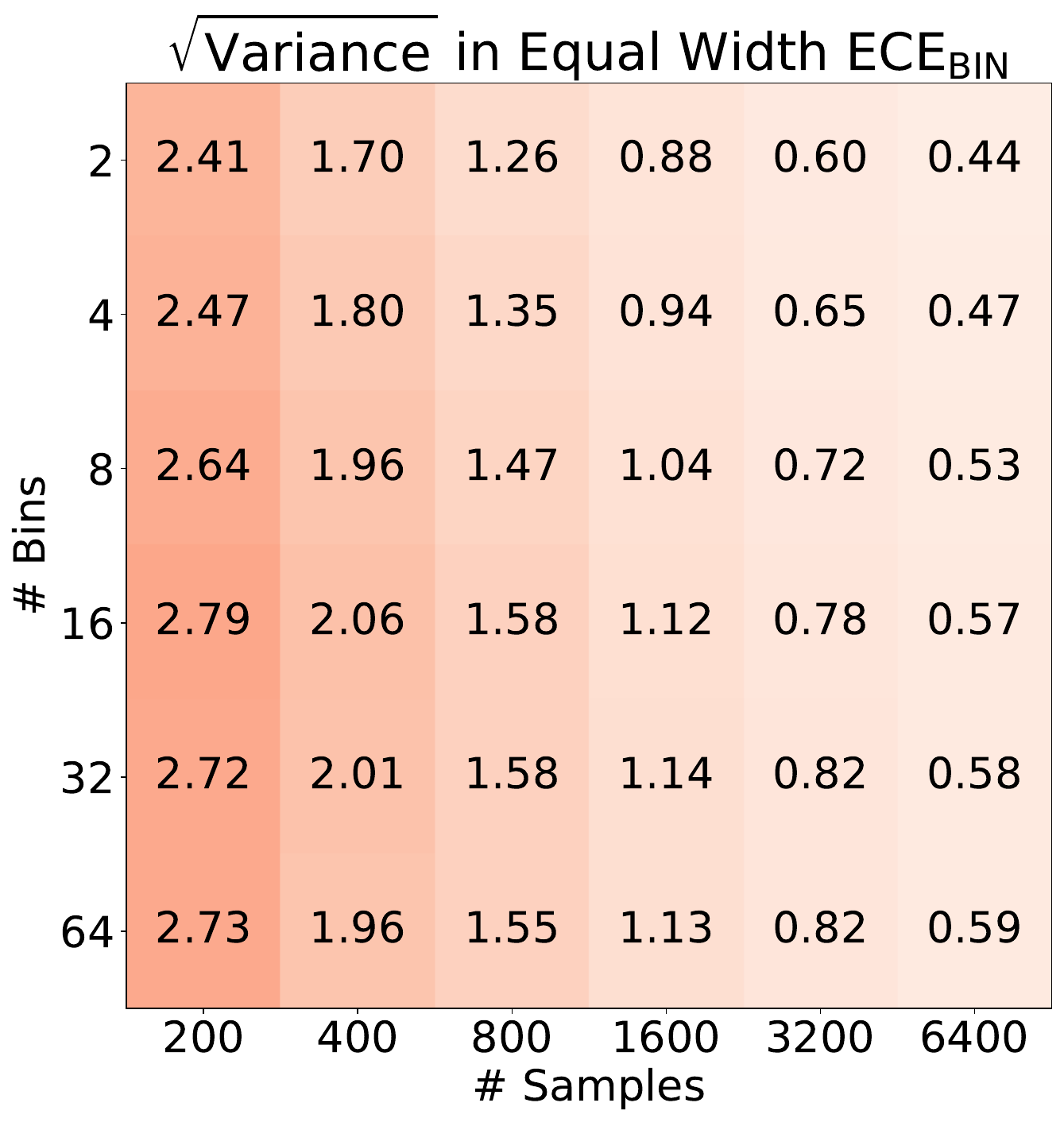}
    \end{subfigure}
    \begin{subfigure}{0.25\textwidth}
      \includegraphics[width=0.9\linewidth]{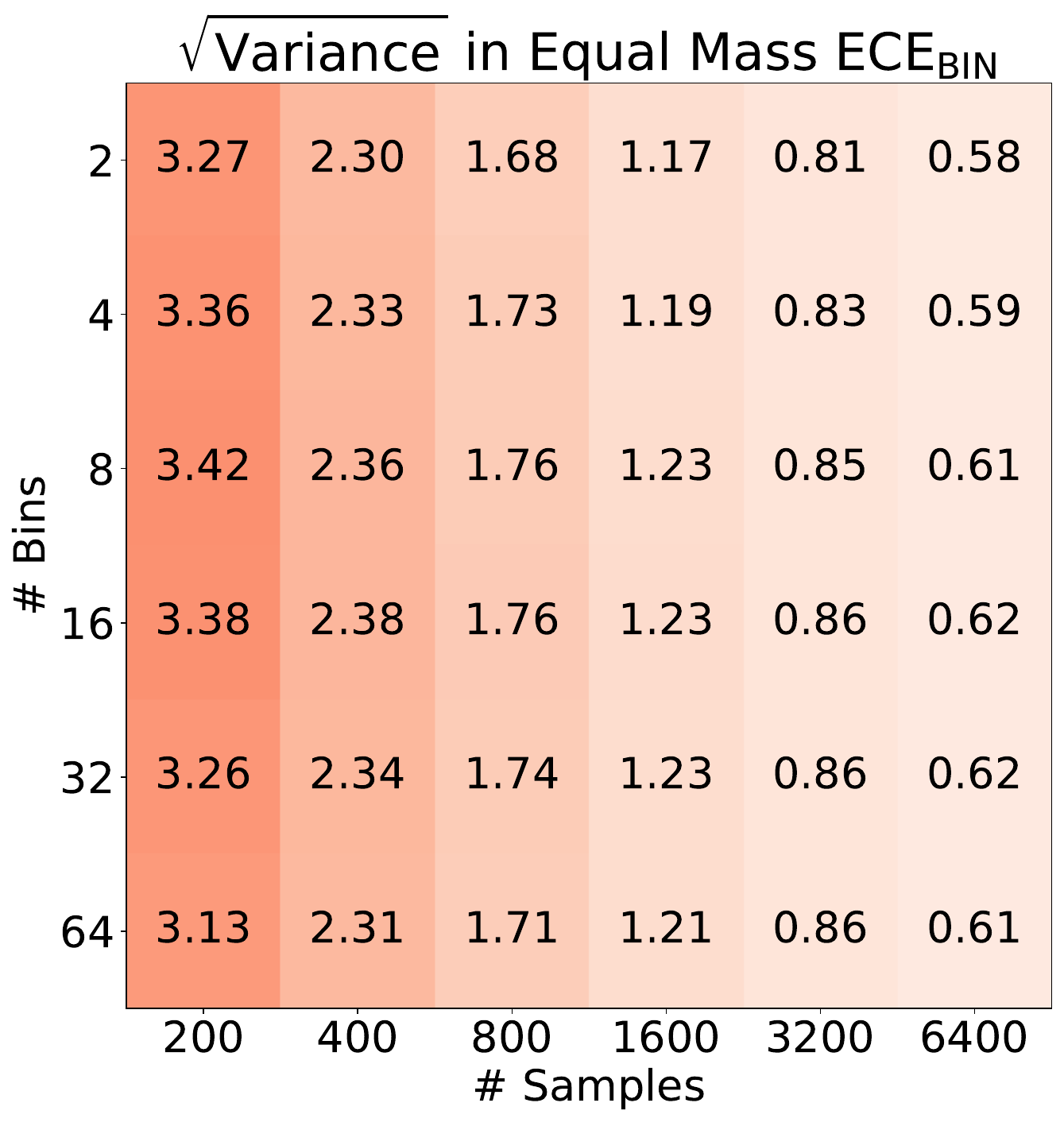}
    \end{subfigure}
    \begin{subfigure}{0.25\textwidth}
      \includegraphics[width=0.9\linewidth]{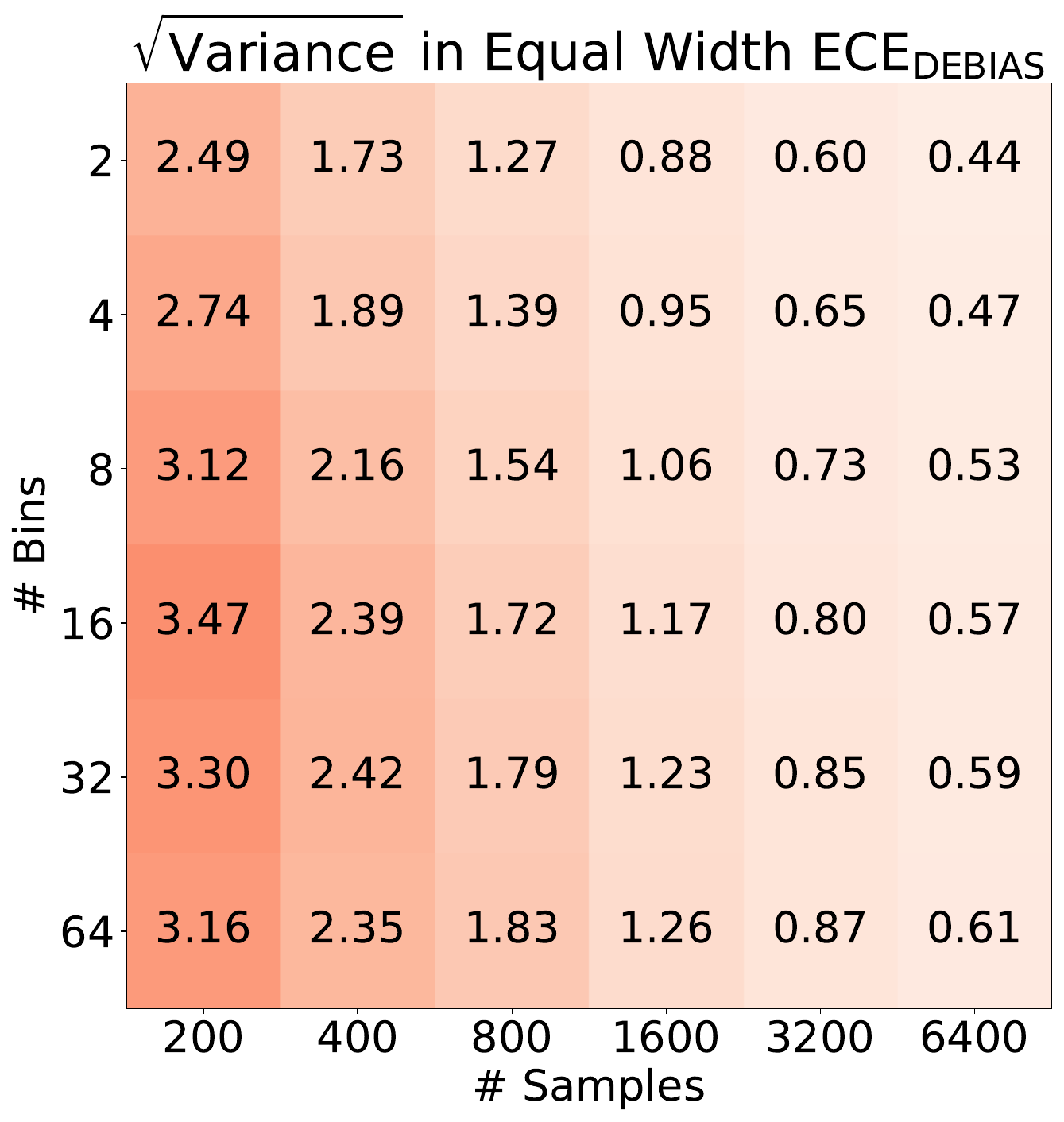}
    \end{subfigure}
    \begin{subfigure}{0.25\textwidth}
      \includegraphics[width=0.9\linewidth]{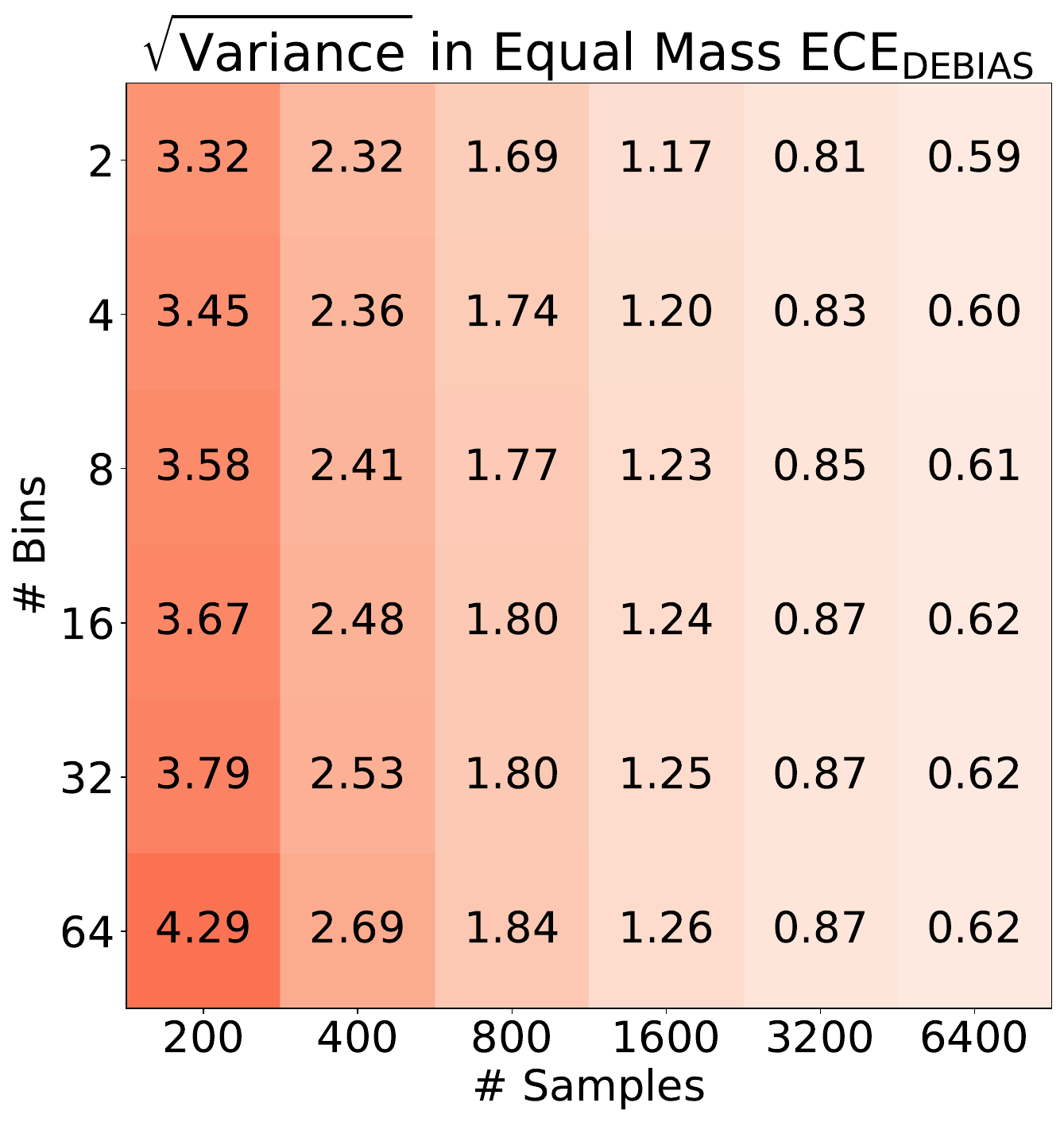}
    \end{subfigure}
    \begin{subfigure}{0.25\textwidth}
      \includegraphics[width=0.9\linewidth]{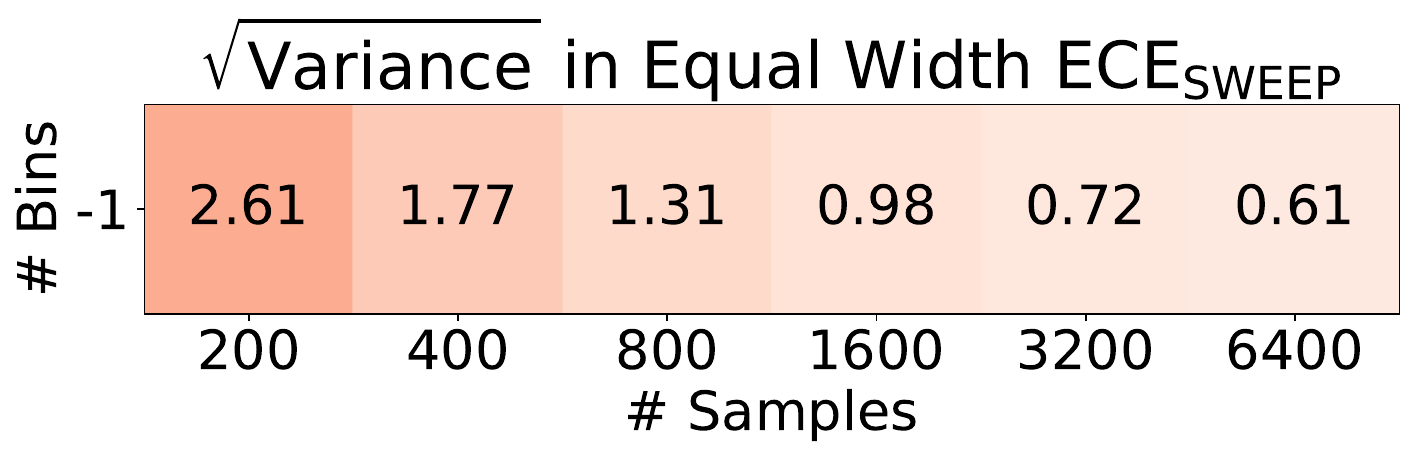}
    \end{subfigure}
    \begin{subfigure}{0.25\textwidth}
      \includegraphics[width=0.9\linewidth]{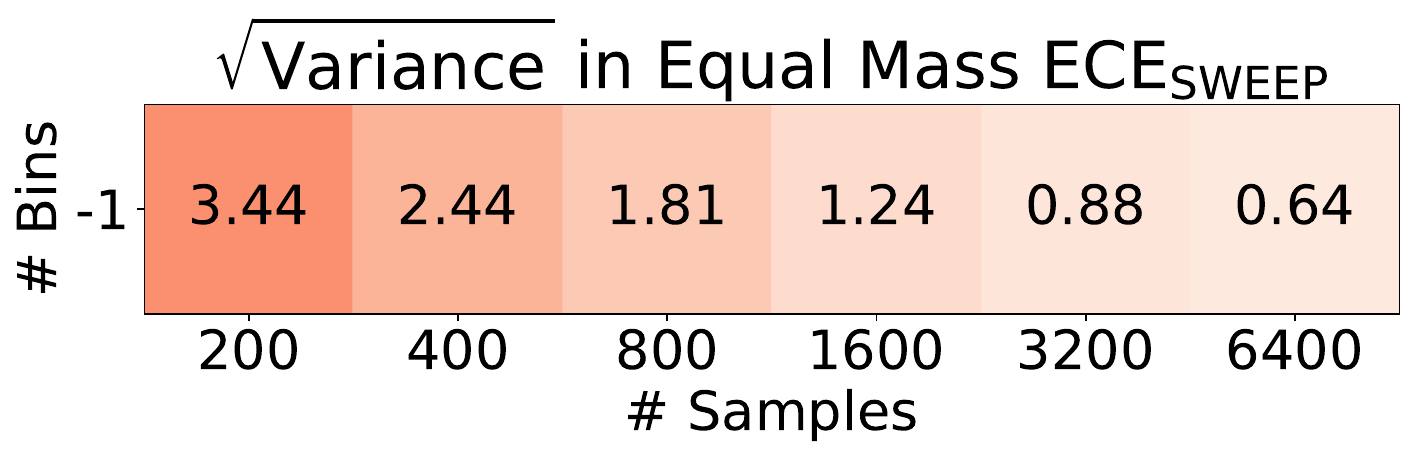}
    \end{subfigure}
    
  \caption{\textbf{$\sqrt{\text{Variance}}$ for various calibration metrics assuming curves fit to CIFAR-100 Wide ResNet-32 output.}
  We plot $\sqrt{\text{Variance}}$ for various calibration metrics using both equal-width binning (left column) and equal-mass binning (right column) as we vary both the sample size $n$ and the number of bins $b$. 
  }
  \label{fig:variance_resnet_wide32_c100}
\end{figure}

\textbf{CIFAR-100 Wide ResNet-32.}
Figure \ref{fig:variance_resnet_wide32_c100} assume parametric curves for $p(f(x))$ and $\truecurve$ that we obtain from maximum-likelihood fits to CIFAR-100 Wide ResNet-32 model output.  

\begin{figure}[htp!]
    \centering
    \begin{subfigure}{0.25\textwidth}
      \includegraphics[width=0.9\linewidth]{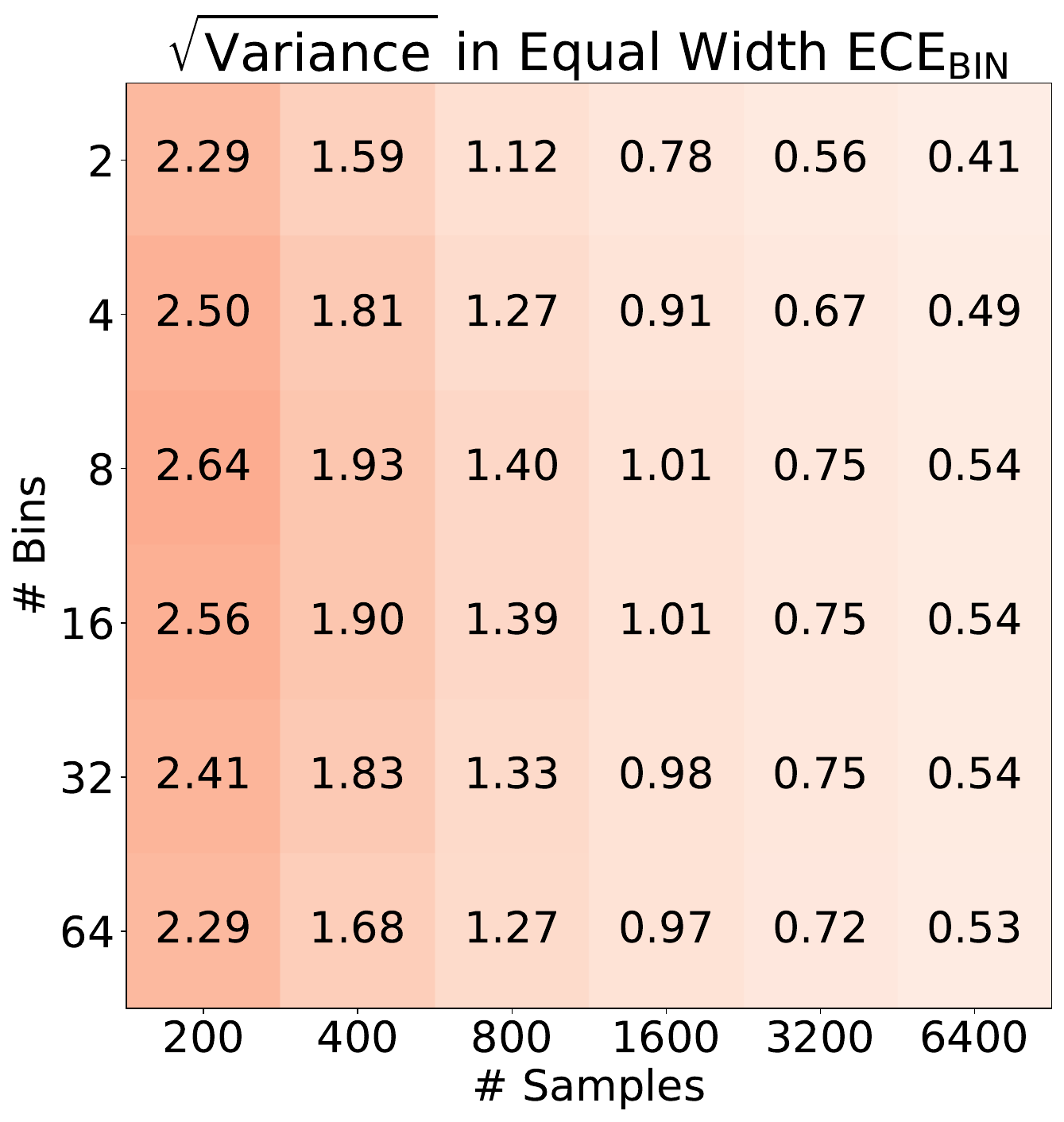}
    \end{subfigure}
    \begin{subfigure}{0.25\textwidth}
      \includegraphics[width=0.9\linewidth]{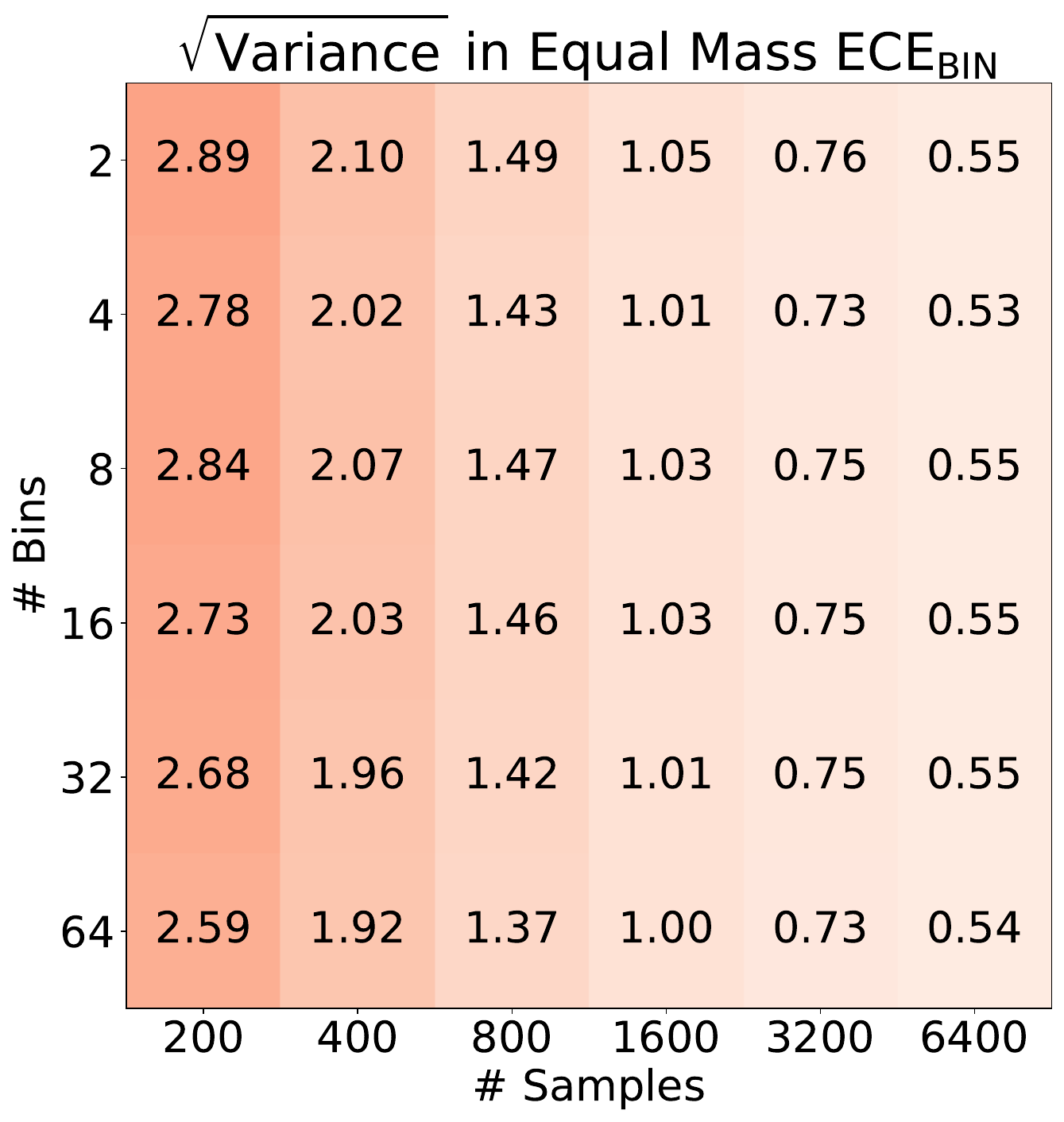}
    \end{subfigure}
    \begin{subfigure}{0.25\textwidth}
      \includegraphics[width=0.9\linewidth]{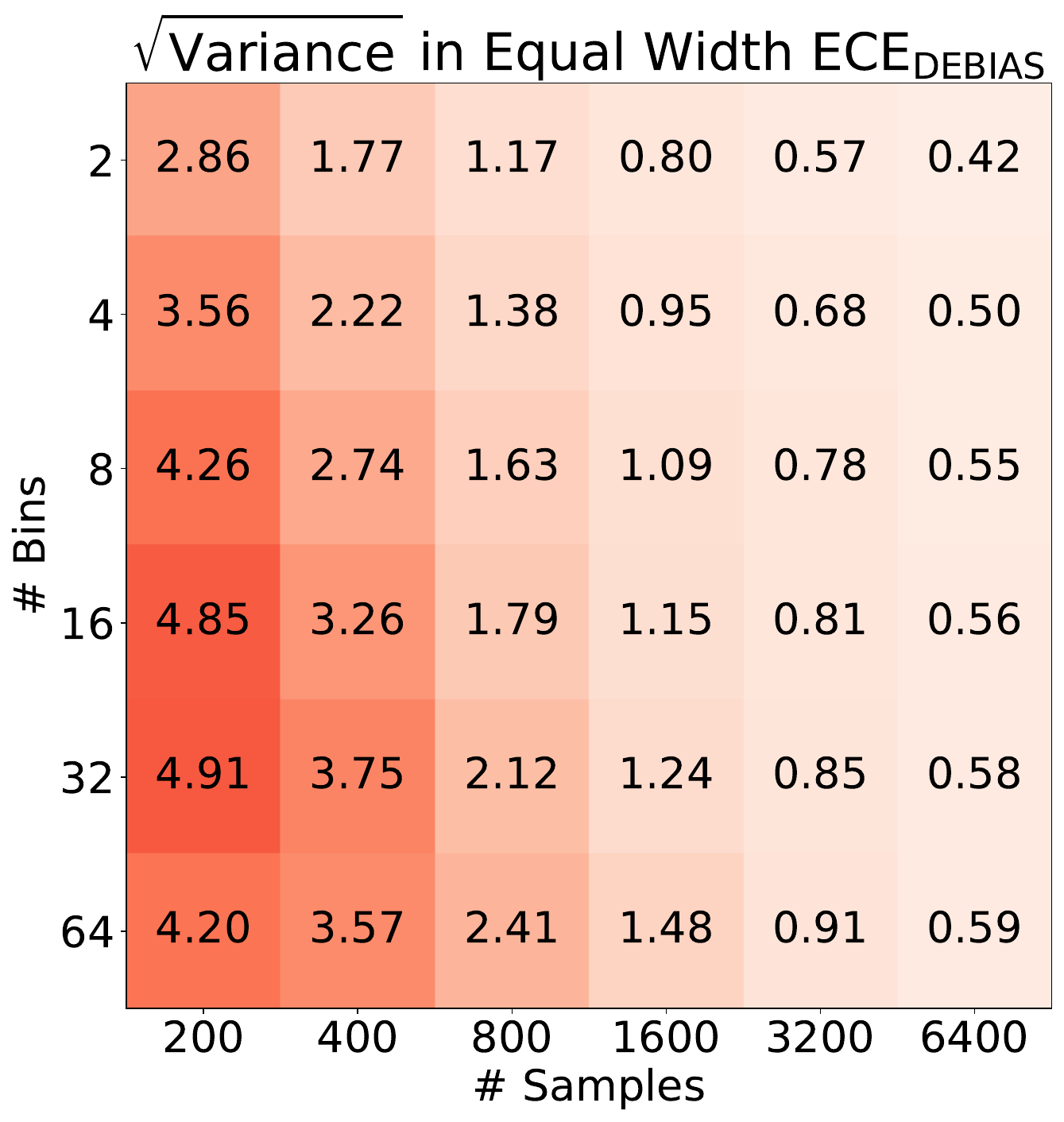}
    \end{subfigure}
    \begin{subfigure}{0.25\textwidth}
      \includegraphics[width=0.9\linewidth]{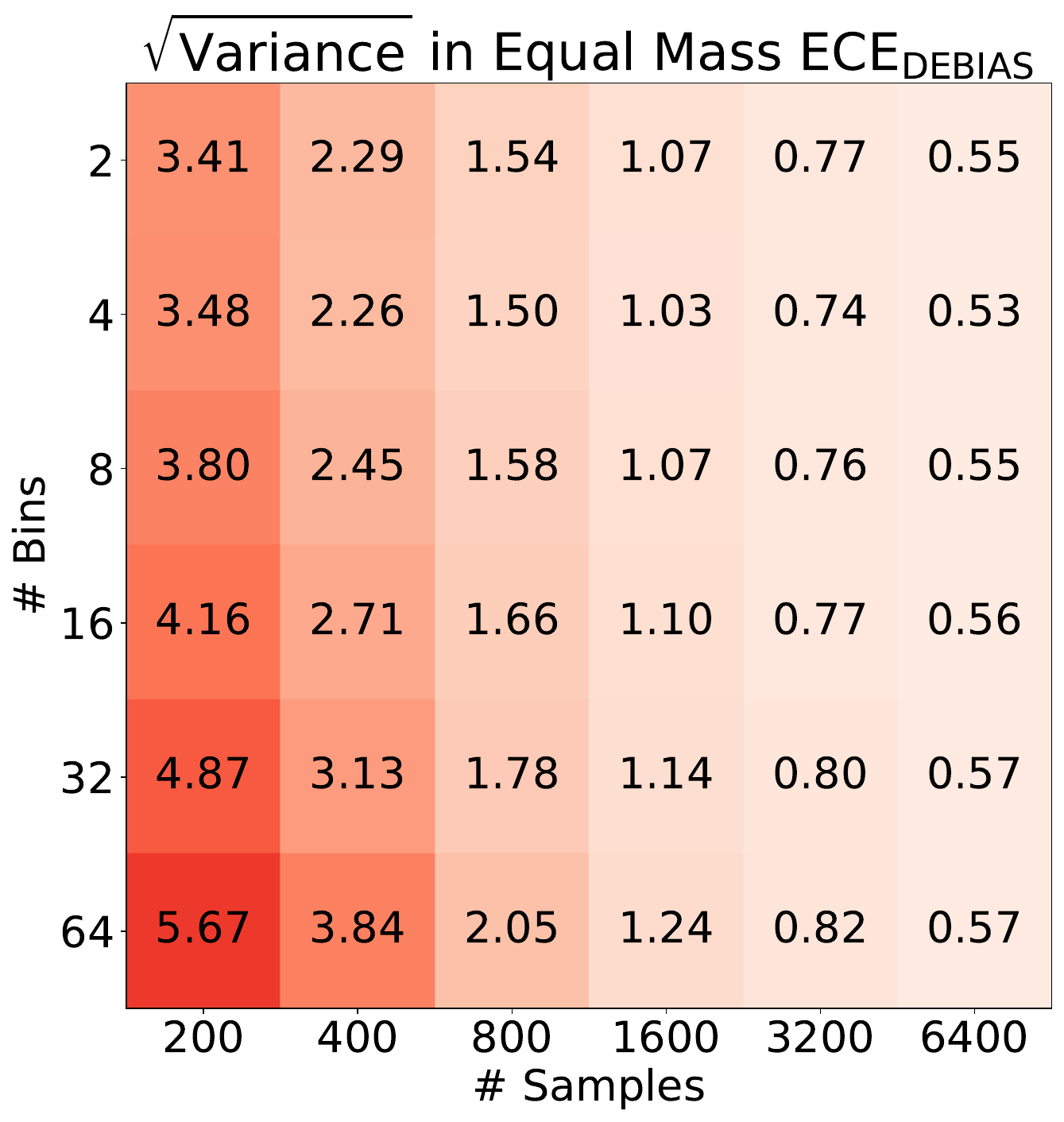}
    \end{subfigure}
    \begin{subfigure}{0.25\textwidth}
      \includegraphics[width=0.9\linewidth]{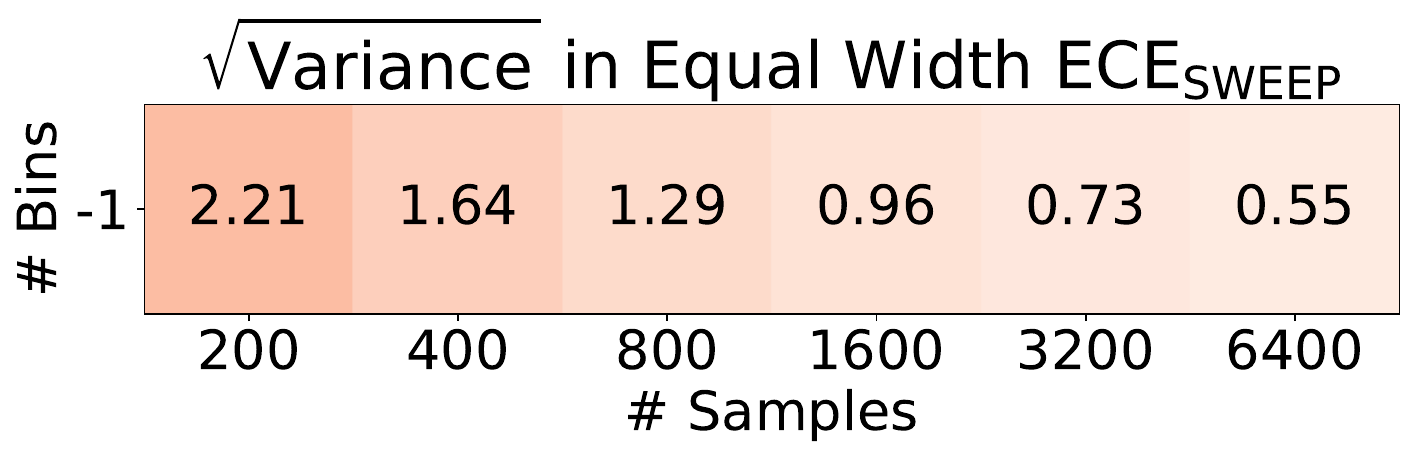}
    \end{subfigure}
    \begin{subfigure}{0.25\textwidth}
      \includegraphics[width=0.9\linewidth]{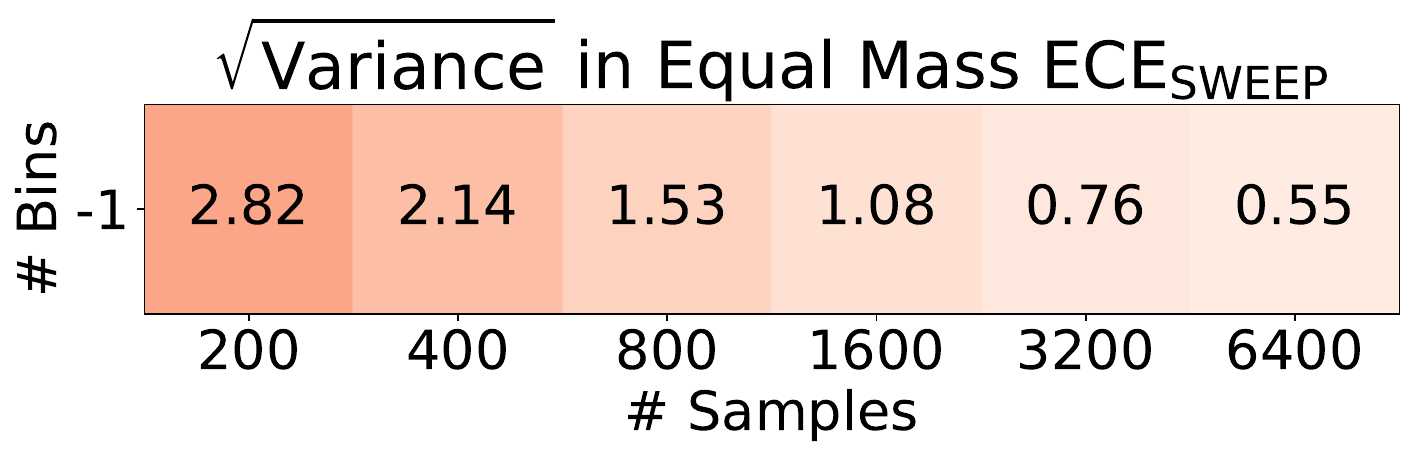}
    \end{subfigure}
    
  \caption{\textbf{$\sqrt{\text{Variance}}$ for various calibration metrics assuming curves fit to ImageNet ResNet-152 output.}
  We plot $\sqrt{\text{Variance}}$ for various calibration metrics using both equal-width binning (left column) and equal-mass binning (right column) as we vary both the sample size $n$ and the number of bins $b$. 
  }
  \label{fig:variance_resnet152_imgnet}
\end{figure}

\textbf{ImageNet ResNet-152.}
Figure \ref{fig:variance_resnet152_imgnet} assume parametric curves for $p(f(x))$ and $\truecurve$ that we obtain from maximum-likelihood fits to ImageNet ResNet-152 model output.  

\clearpage
\newpage

\section{Controlling true calibration error using BBC}
\label{apx:controlling_tce}
\begin{figure}[b!]
  \centering
   \includegraphics[width=0.9\linewidth]{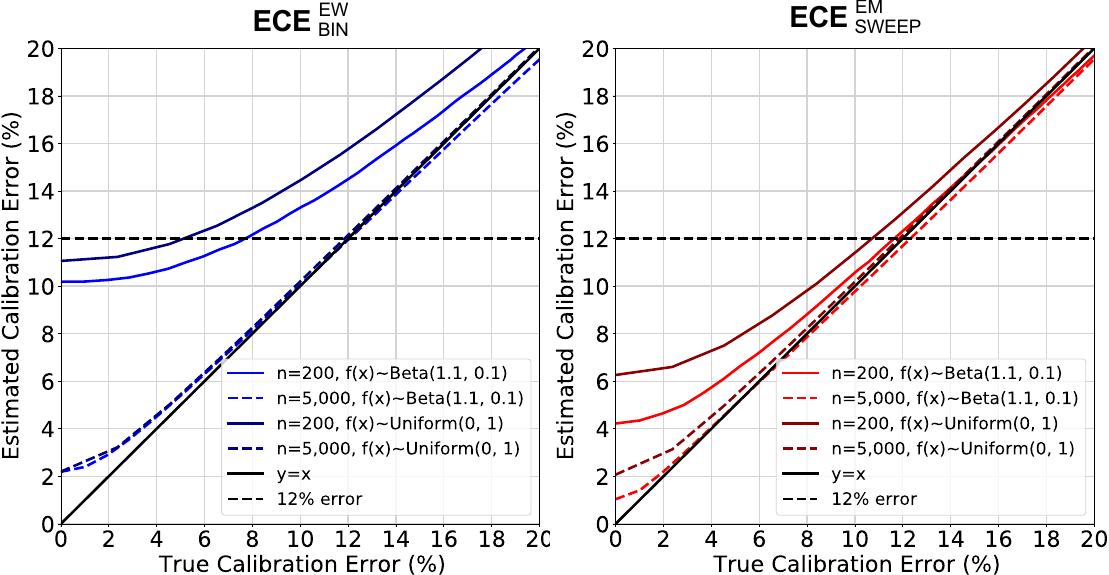}
  \caption{Bias in calibration estimation increases as \tce{} decreases. Average ECE (\%) for \ewece{} (left) and \emmonsweep{} (right) versus the \tce{} (\%), with varying sample size and score distributions. The estimator bias is systematically worse for better calibrated models, and the effect is more egregious with fewer samples. At $n=200$ samples, depending on the score distribution, an \ewece{} estimate of 12\% could either correspond to 5\% or 8\% \tce{}.  \emmonsweep{} somewhat mitigates the bias and ambiguity in calibration error estimation.}
  \label{fig:bias_vs_true_ce}
\end{figure}

We evaluate the estimation bias of calibration estimators as we systematically vary the \tce{}.      Figure \ref{fig:bias_vs_true_ce}  shows the average estimated calibration error for \ewece{} and \emmonsweep{} versus the \tce{}. The average calibration error is computed across $m=1{,}000$ simulated datasets, and we include results for two sample sizes, $n=200$ and $n=5{,}000$, and two score distributions, $f(x) \sim \Uniform(0,1)$ and $f(x) \sim \Beta(1.1, 0.1)$, the beta distribution fit to the CIFAR-100 Wide ResNet\_32.  To control the \tce{}, we assume $\truecurve = c^d$ and vary $d \in [1,10]$. When $d=1$ the true calibration curve is $\truecurve=c$, which means the model's predicted confidence score is exactly equal to its empirical accuracy and thus the \tce{} is 0\%.  As we increase $d$, we move the true calibration curve farther away from the perfect calibration curve, which increases the \tce{} of the model. 

The estimation bias can be seen visually as the difference between the ECE and the $y=x$ line. Perfect estimation (0 bias) corresponds to the $y=x$ line.  Bias is highest when the model is perfectly calibrated (\tce{} is 0\%) and generally decreases as \tce{} increases. A larger sample size of $n=5,000$ reduces the bias, but with perfectly calibration \ecebin{} can still be off by 2\%.  The \emmonsweep{} metric significantly reduces this bias. 

\section{What number of bins does \texorpdfstring{\emmonsweep}{ECEsweep} choose?}
\label{apx:chosen_bins}

\begin{figure}[ht!] 
  \centering
  \begin{subfigure}{0.48\textwidth}
    \centering
    \includegraphics[width=0.8\linewidth, height=0.8\linewidth]{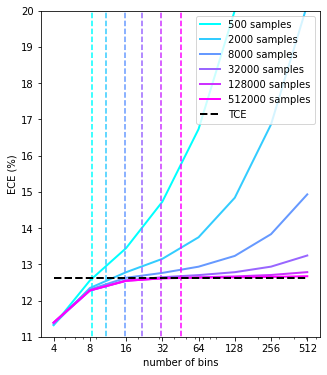}
    \caption{Uncalibrated model.}
  \end{subfigure}
  \begin{subfigure}{0.48\textwidth}
    \centering
    \includegraphics[width=0.8\linewidth, height=0.8\linewidth]{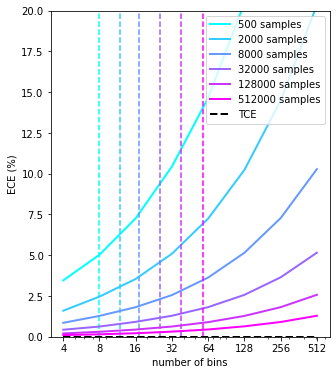}
    \caption{Perfectly calibrated model.}
  \end{subfigure}
  \caption{\textbf{Bins chosen by equal mass \cesweep{} method}. We plot equal mass \ecebin{} \% versus number of bins for various sample sizes $n$.  We highlight the TCE with a horizontal dashed line and show the average number of bins chosen by the \cesweep{} method for different sample sizes with vertical dashed lines.  When the model is uncalibrated (left) \cesweep{} chooses a bin number that is close to optimal.  However, for perfectly calibrated models (right), the optimal number of bins is small (<=4), and \cesweep{} does not do a good job of selecting a good bin number.  The incorrect bin selection may partially explain why \cesweep{} still has some bias for perfectly calibrated models. However, we note that any binning-based technique that always outputs a positive number will never be completely unbiased for perfectly calibrated models.
  }
  \label{fig:bins_chosen}
\end{figure}

For Figure \ref{fig:bins_chosen}, the uncalibrated plot assumes $\truecurve = \text{logistic}(10*c-5)$ while the calibrated plot assumes $\truecurve = c$.  Both experiments assume $f(x) \sim \text{Uniform}(0,1)$.

\end{document}